\documentclass[10pt,twocolumn,letterpaper]{article}
\usepackage{cvpr}
\usepackage{times}
\usepackage{epsfig}
\usepackage{graphicx}
\usepackage{amsmath}
\usepackage{amssymb}
\usepackage{rotating}
\usepackage{color}
\usepackage{framed}
\usepackage{subcaption}
\usepackage{booktabs}
\usepackage{bm}
\usepackage{bbm}
\usepackage{enumitem}
\usepackage{tikz}
\usepackage{tikz-dependency}
\usepackage[export]{adjustbox}
\usetikzlibrary{%
  shapes,%
  arrows,%
  positioning,%
  calc,%
  automata%
}

\usepackage[pagebackref=true,breaklinks=true,letterpaper=true,colorlinks,bookmarks=false]{hyperref}

\cvprfinalcopy

\definecolor{pf7}{RGB}{166, 118, 29}

\usepackage{amsmath}

\def \ie {\emph{i.e.}}

\def \etal {\emph{et al.}}

\def \ours {Ours\xspace}
\newcommand{\tit}[1]{\smallbreak\noindent\textbf{#1.}}
\newcommand{\tinytit}[1]{\noindent\textbf{#1.}}

\begin{document}

\title{\textit{Show, Control and Tell}:\\A Framework for Generating Controllable and Grounded Captions}

\author{Marcella Cornia, Lorenzo Baraldi, Rita Cucchiara \\
University of Modena and Reggio Emilia \\
{\tt\small \{name.surname\}@unimore.it}
}

\maketitle

\begin{abstract}
Current captioning approaches can describe images using black-box architectures whose behavior is hardly controllable and explainable from the exterior. As an image can be described in infinite ways depending on the goal and the context at hand, a higher degree of controllability is needed to apply captioning algorithms in complex scenarios.
In this paper, we introduce a novel framework for image captioning which can generate diverse descriptions by allowing both grounding and controllability. Given a control signal in the form of a sequence or set of image regions, we generate the corresponding caption through a recurrent architecture which predicts textual chunks explicitly grounded on regions, following the constraints of the given control.
Experiments are conducted on Flickr30k Entities and on COCO Entities, an extended version of COCO in which we add grounding annotations collected in a semi-automatic manner. Results demonstrate that our method achieves state of the art performances on controllable image captioning, in terms of caption quality and diversity.
Code and annotations are publicly available at: \url{https://github.com/aimagelab/show-control-and-tell}.
\end{abstract}

\section{Introduction}
\label{sec:introduction}


Image captioning brings vision and language together in a generative way. As a fundamental step towards machine intelligence, this task has been recently gaining much attention thanks to the spread of Deep Learning architectures which can effectively describe images in natural language~\cite{vinyals2015show,karpathy2015deep,xu2015show,vinyals2017show}. Image captioning approaches are usually capable of learning a correspondence between an input image and a probability distribution over time, from which captions can be sampled either using a greedy decoding strategy~\cite{vinyals2017show}, or more sophisticated techniques like beam search and its variants~\cite{anderson2016guided}.

As the two main components of captioning architectures are the image encoding stage and the language model, researchers have focused on improving both phases, which resulted in the emergence of attentive models~\cite{xu2015show} on one side, and of more sophisticated interactions with the language model on the other~\cite{lu2018neural,baraldi2017hierarchical}. Recently, attentive models have been improved by replacing the attention over a grid of features with attention over image regions~\cite{anderson2018bottom,wang2018object,zhang2018interpretable}. In these models, the generative process attends a set of regions which are softly selected while generating the caption.

\begin{figure}[t]
    \centering
    \includegraphics[width=\linewidth]{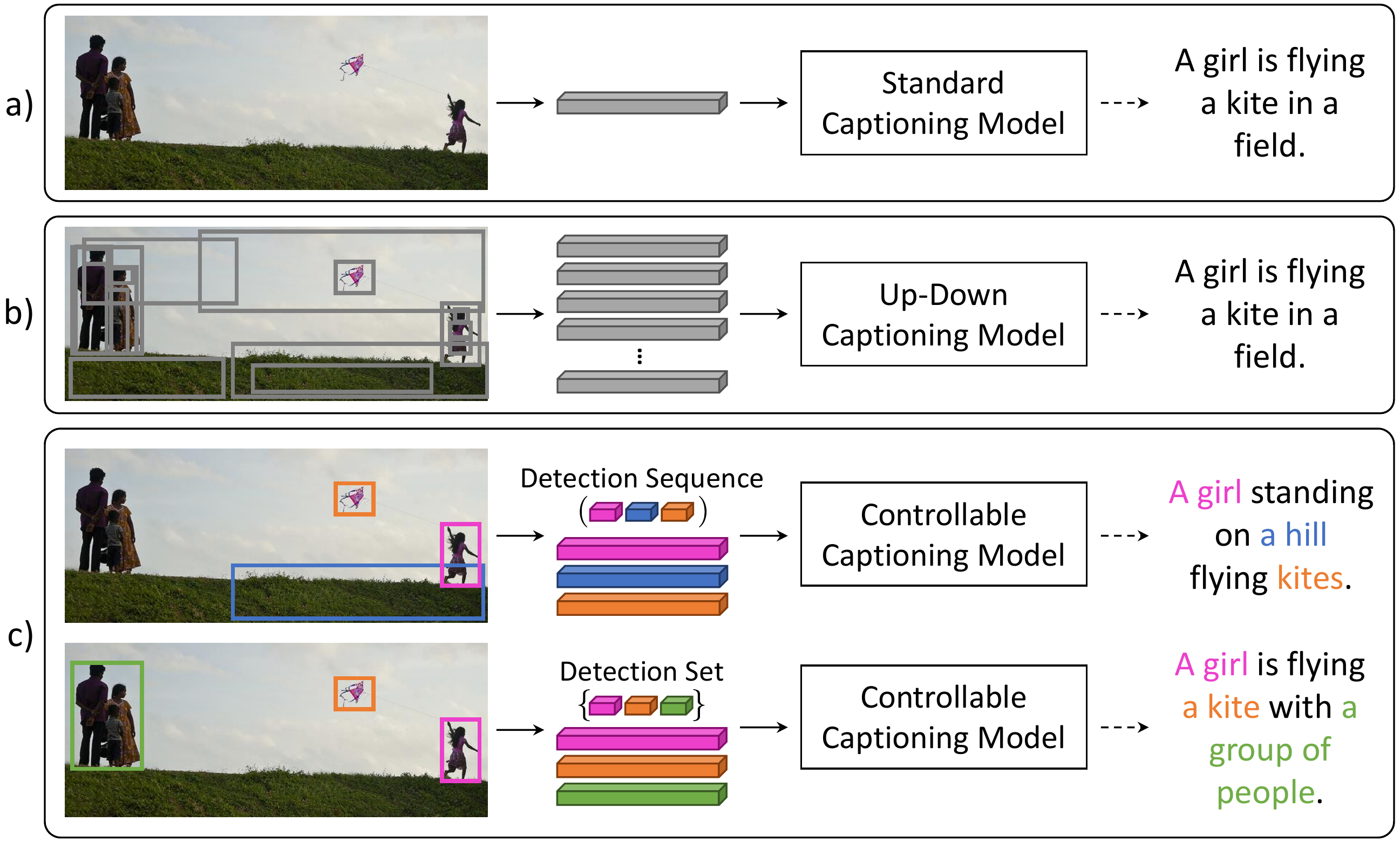}
    \caption{Comparison between (a) captioning models with global visual feature~\cite{vinyals2017show}, (b) attentive models which integrate features from image regions~\cite{anderson2018bottom} and (c) our \textit{Show, Control and Tell}. Our method can produce multiple captions for a given image, depending on a control signal which can be either a sequence or a set of image regions. Moreover, chunks of the generated sentences are explicitly grounded on regions.}
    \label{fig:first_page}
    \vspace{-.3cm}
\end{figure}

Despite these advancements, captioning models still lack controllability and explainability -- \ie, their behavior can hardly be influenced and explained. As an example, in the case of attention-driven models, the architecture implicitly selects which regions to focus on at each timestep, but it cannot be supervised from the exterior. While an image can be described in multiple ways, such an architecture provides no way of controlling which regions are described and what importance is given to each region. This lack of controllability creates a distance between human and machine intelligence, as humans can manage the variety of ways in which an image can be described, and select the most appropriate one depending on the task and the context at hand. Most importantly, this also limits the applicability of captioning algorithms to complex scenarios in which some control over the generation process is needed. As an example, a captioning-based driver assistance system would need to focus on dangerous objects on the road to alert the driver, rather than describing the presence of trees and cars when a risky situation is detected. Eventually, such systems would also need to be explainable, so that their behavior could be easily interpreted in case of failures. 

In this paper, we introduce \textit{Show, Control and Tell}, that explicitly addresses these shortcomings (Fig.~\ref{fig:first_page}). It can generate diverse natural language captions depending on a control signal which can be given either as a sequence or as a set of image regions which need to be described. As such, our method is capable of describing the same image by focusing on different regions and in a different order, following the given conditioning. 
Our model is built on a recurrent architecture which considers the decomposition of a sentence into noun chunks and models the relationship between image regions and textual chunks, so that the generation process can be explicitly grounded on image regions. To the best of our knowledge, this is the first captioning framework controllable from image regions.

\tit{Contributions} Our contributions are as follows:
\begin{itemize}[noitemsep,topsep=0pt]
    \item We propose a novel framework for image captioning which is controllable from the exterior, and which can produce natural language captions explicitly grounded on a sequence or a set of image regions.
    \item The model explicitly considers the hierarchical structure of a sentence by predicting a sequence of noun chunks. Also, it takes into account the distinction between visual and textual words, thus providing an additional grounding at the word level.
    \item We evaluate the model with respect to a set of carefully designed baselines, on Flickr30k Entities and on COCO, which we semi-automatically augment with grounding image regions for training and evaluation purposes.
    \item  Our proposed method achieves state of the art results for controllable image captioning on Flick30k and COCO both in terms of diversity and caption quality, even when compared with methods which focus on diversity.
\end{itemize}


\section{Related work}
\label{sec:related}

A large number of models has been proposed for image captioning~\cite{rennie2017self,yang2016review,lu2017knowing,liu2017improved,johnson2016densecap,lu2018neural}. Generally, all integrate recurrent neural networks as language models, and a representation of the image which might be given by the output of one or more layer of a CNN~\cite{vinyals2017show,donahue2015long,rennie2017self,lu2017knowing}, or by a time-varying vector extracted with an attention mechanism~\cite{xu2015show,you2016image,lu2017knowing,cornia2018paying,anderson2018bottom} selected either from a grid over CNN features, or integrating image regions eventually extracted from a detector~\cite{pedersoli2016areas,anderson2018bottom}. Attentive models provided a first way of grounding words to parts of the image, although with a blurry indication which was rarely semantically significant. Regarding the training strategies, notable advances have been made by using Reinforcement Learning to train non-differentiable captioning metrics~\cite{ranzato2015sequence,liu2017improved,rennie2017self}. In this work, we propose an extended version of this approach which deals with multiple output distributions and rewards the alignment of the caption to the control signal.

Recently, more principled approaches have been proposed for grounding a caption on the image~\cite{plummer2015flickr30k,rohrbach2016grounding,hu2017modeling,hu2016natural}: DenseCap~\cite{johnson2016densecap} generates descriptions for specific image regions. Further, the Neural Baby Talk approach~\cite{lu2018neural} extends the attentive model in a two-step design in which a word-level sentence template is firstly generated and then filled by object detectors with concepts found in the image. We instead decompose the caption at the level of noun chunks, and explicitly ground each of them to a region. This approach has the additional benefit of providing an explicability method at the chunk level. 

Another related line of work is that of generating diverse descriptions. Some works have extended the beam-search algorithm to sample multiple captions from the same distribution~\cite{vijayakumar2018diverse,anderson2016guided}, while different GAN-based approaches have also appeared~\cite{dai2017towards,shetty2017speaking,wang2017diverse}. Most of these improve on diversity, but suffer on accuracy and do not provide controllability over the generation process. Others have conditioned the generation with a specific style or sentiment~\cite{mathews2018semstyle,mathews2016senticap,gan2017stylenet}. Our work is mostly related to~\cite{deshpande2018diverse}, which uses a control input as a sequence of part-of-speech tags. This approach, while generating diversity, is hardly employable to effectively control the generation of the sentence; in contrast, we use image regions as a controllability method.

\section{Method}
\label{sec:approach}

\begin{figure}[t]
    \centering
    \begin{tabular}{c}
        \includegraphics[width=0.32\linewidth]{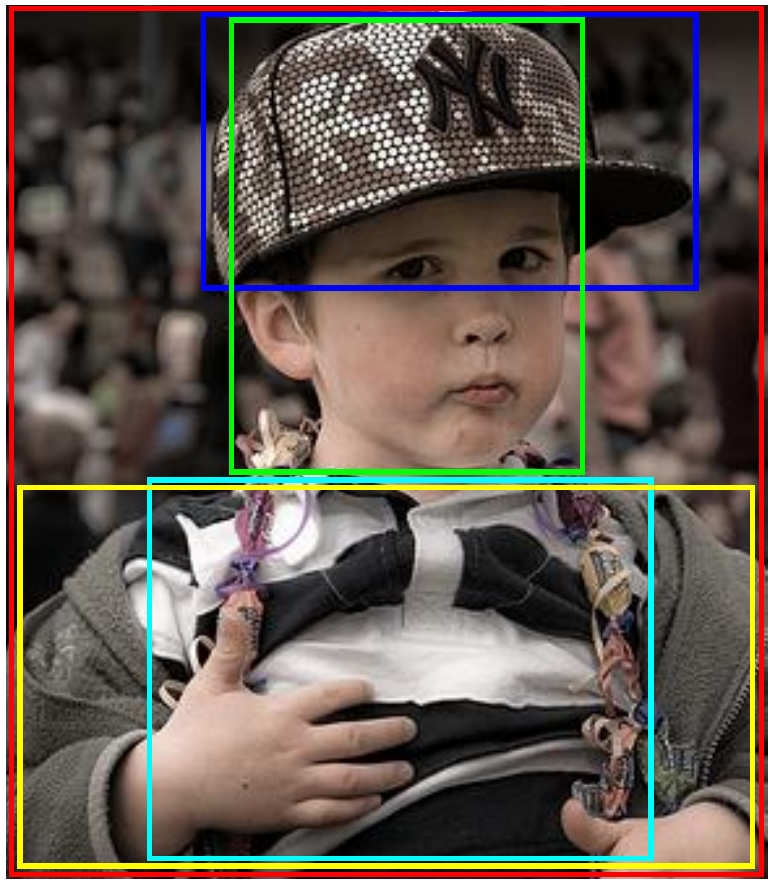}
        \resizebox{0.65\linewidth}{!}{
            \begin{tikzpicture}[
                level/.style={sibling distance=2cm,
                level distance = 1cm},
                level 1/.style={sibling distance=4cm}
            ]
                \begin{scope}[shift={(1cm,0in)}]
                \begin{deptext}[column sep=.3em]
                    \textbf{A} \& \textbf{young} \& \textbf{boy} \& \textbf{with} \& \textbf{a} \& \textbf{cap} \& \textbf{on} \&
                    \textbf{his} \& \textbf{head,} \\
                \end{deptext}
            
                \depedge[edge above]{3}{1}{\bf\textcolor{pf7}{det}}
                \depedge[edge above]{3}{2}{\bf\textcolor{pf7}{amod}}
                             \depedge[edge above]{3}{4}{\bf\textcolor{pf7}{prep}}        
                \depedge[edge above]{4}{6}{\bf\textcolor{pf7}{pobj}}
                \depedge[edge above]{6}{5}{\bf\textcolor{pf7}{det}}  
                \depedge[edge above]{6}{7}{\bf\textcolor{pf7}{prep}}
                \depedge[edge above]{7}{9}{\bf\textcolor{pf7}{pobj}}
                \depedge[edge above]{9}{8}{\bf\textcolor{pf7}{poss}}
                \wordgroup[group style={fill=red!40, draw=red}]{1}{1}{3}{G1}
                \wordgroup[group style={fill=blue!40, draw=blue}]{1}{5}{6}{G1}
                \wordgroup[group style={fill=green!40, draw=green}]{1}{8}{9}{G2}
                
                \end{scope}
                
                \begin{scope}[shift={(1cm,-1cm)}]
                \begin{deptext}[column sep=.3em]
                    \textbf{striped} \& \textbf{shirt,} \& \textbf{and} \& \textbf{gray} \&
                    \textbf{sweat} \& \textbf{jacket.} \\
                \end{deptext}
                \end{scope}

                \depedge[edge below]{2}{1}{\bf\textcolor{pf7}{amod}}
                \depedge[edge below]{2}{3}{\bf\textcolor{pf7}{cc}}
                \depedge[edge below]{6}{4}{\bf\textcolor{pf7}{amod}}
                \depedge[edge below]{6}{5}{\bf\textcolor{pf7}{compound}}
                \wordgroup[group style={fill=cyan!40, draw=cyan}]{1}{1}{2}{G2}
                \wordgroup[group style={fill=yellow!40, draw=yellow}]{1}{4}{6}{G2}
                
            \end{tikzpicture}
        }
        \end{tabular}
    \caption{Example of a dependency tree for a caption. Noun chunks are marked with rounded boxes; chunks corresponding to image regions are depicted using the same color.}
    \label{fig:example_nlp}
    \vspace{-.3cm}
\end{figure}

Sentences are natural language structures which are hierarchical by nature~\cite{manning1999foundations}. At the lowest level, a sentence might be thought as a sequence of words: in the case of a sentence describing an image, we can further distinguish between \textit{visual} words, which describe something visually present in the image, and \textit{textual} words, that refer to entities which are not present in the image~\cite{lu2018neural}.
Analyzing further the syntactic dependencies between words, we can recover a higher abstraction level in which words can be organized into a tree-like structure: in a dependency tree~\cite{goldberg2013training,honnibal2013non,chen2014fast}, each word is linked together with its modifiers (Fig.~\ref{fig:example_nlp}).

\begin{figure*}[t]
    \centering
    \includegraphics[width=0.98\linewidth]{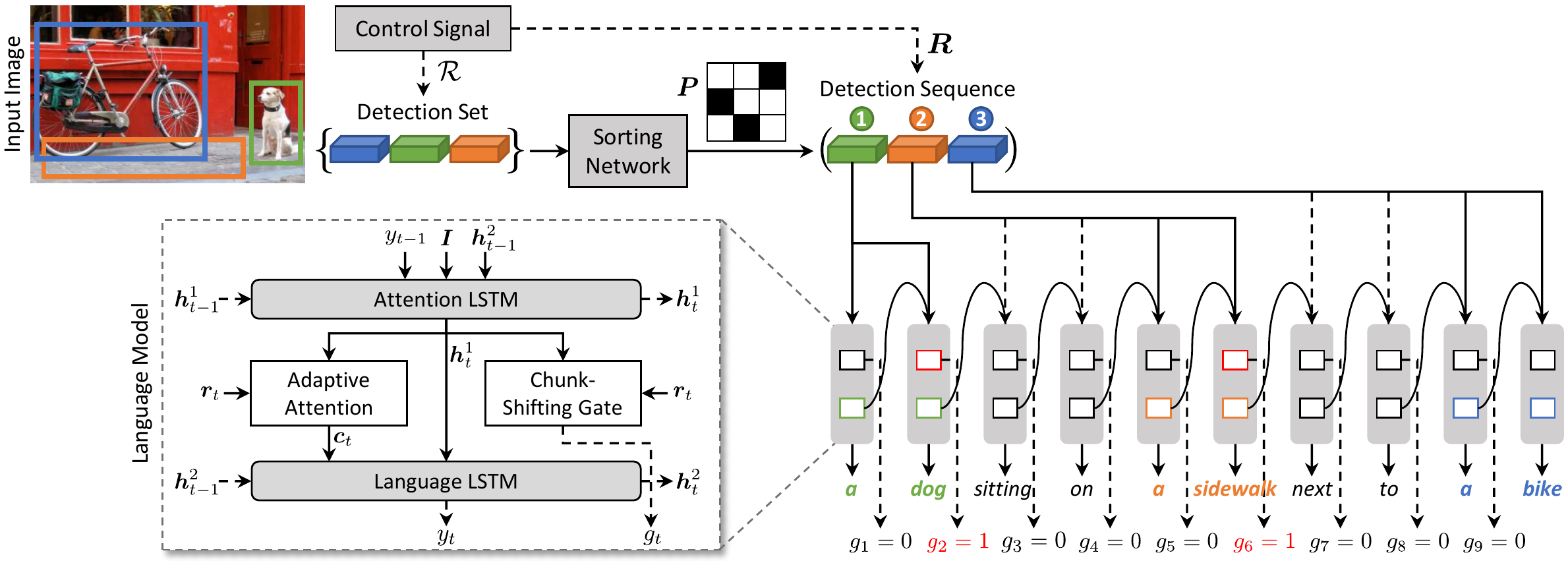}
    \caption{Overview of the approach. Given an image and a control signal, the figure shows the process to generate the controlled caption and the architecture of the language model.}
    \label{fig:model}
    \vspace{-.3cm}
\end{figure*}

Given a dependency tree, nouns can be grouped with their modifiers, thus building \textit{noun chunks}. For instance, the caption depicted in Fig.~\ref{fig:example_nlp} can be decomposed into a sequence of different noun chunks: ``a young boy'', ``a cap'', ``his head'', ``striped shirt'', and ``gray and sweat jacket''.
As noun chunks, just like words, can be visually grounded into image regions, a caption can also be mapped to a sequence of regions, each corresponding to a noun chunk. A chunk might also be associated with multiple image regions of the same class if more than one possible mapping exists.

The number of ways in which an image can be described results in different sequences of chunks, linked together to form a fluent sentence. Therefore, captions also differ in terms of the set of considered regions, the order in which they are described, and their mapping to chunks given by the linguistic abilities of the annotator.

Following these premises, we define a model which can recover the variety of ways in which an image can be described, given a control input expressed as a sequence or set of image regions. We begin by presenting the former case, and then show how our model deals with the latter scenario.

\subsection{Generating controllable captions}
Given an image $\bm{I}$ and an ordered sequence of set of regions $\bm{R} = ( \bm{r}_0, \bm{r}_1, ..., \bm{r}_N)$\footnote{For generality, we will always consider sequences of sets of regions, to deal with the case in which a chunk in the target sentence can be associated to multiple regions in training and evaluation data.}, the goal of our captioning model is to generate a sentence $\bm{y} = \left( y_0, y_1, ..., y_T\right)$ which in turns describes all the regions in $\bm{R}$ while maintaining the fluency of language.

Our model is conditioned on both the input image $\bm{I}$ and the sequence of region sets $\bm{R}$, which acts as a control signal, and jointly predicts two output distributions which correspond to the word-level and chunk-level representation of the sentence: the probability of generating a word at a given time, \ie~$p(y_t | \bm{R}, \bm{I}; \bm{\theta})$, and that of switching from one chunk to another, \ie~$p(g_t | \bm{R}, \bm{I}; \bm{\theta})$, where $g_t$ is a boolean chunk-shifting gate. During the generation, the model maintains a pointer to the current region set $\bm{r}_i$ and can shift to the next element in $\bm{R}$ by means of the gate $g_t$.

To generate the output caption, we employ a recurrent neural network with adaptive attention. At each timestep, we compute the hidden state $\bm{h}_t$ according to the previous hidden state $\bm{h}_{t-1}$, the current image region set $\bm{r}_t$ and the current word $w_t$, such that $\bm{h}_t = \text{RNN}(w_t, \bm{r}_t, \bm{h}_{t-1})$. At training time, $\bm{r}_t$ and $w_t$ are the ground-truth region set and word corresponding to timestep $t$; at test time, $w_t$ is sampled from the first distribution predicted by the model, while the choice of the next image region is driven by the values of the chunk-shifting gate sampled from the second distribution:
\begin{equation}
    \label{eq:r_t}
    \bm{r}_{t+1} \leftarrow \bm{R}[i], \quad \text{where~} i = \min\left( \sum_{k=1}^{t}{g_k}, N\right),~ g_k \in \{0,1\}
\end{equation}
where $\{ g_k \}_k$ is the sequence of sampled gate values, and $N$ is the number of region sets in $\bm{R}$.

\tit{Chunk-shifting gate}
We compute $p(g_t | \bm{R})$ via an adaptive mechanism in which the LSTM computes a compatibility function between its internal state and a latent representation which models the state of the memory at the end of a chunk. The compatibility score is compared to that of attending one of the regions in $\bm{r}_t$, and the result is used as an indicator to switch to the next region set in $\bm{R}$.

The LSTM is firstly extended to obtain a chunk sentinel $\bm{s}^c_t$, which models a component extracted from the memory encoding the state of the LSTM at the end of a chunk. The sentinel is computed as:
\begin{align}
    \label{eq:s_g}
    \bm{l}^c_t &= \sigma(\bm{W}_{ig} \bm{x}_t + \bm{W}_{hg} \bm{h}_{t-1}) \\
    \bm{s}^c_t &= \bm{l}^c_t\odot\tanh(\bm{m}_t)
\end{align}
where $\bm{W}_{ig} \in \mathbb{R}^{d\times k}$, $\bm{W}_{hg} \in \mathbb{R}^{d\times d}$ are learnable weights, $\bm{m}_t \in \mathbb{R}^{d}$ is the LSTM cell memory and $\bm{x}_t \in \mathbb{R}^k$ is the input of the LSTM at time $t$; $\odot$ represents the Hadamard element-wise product and $\sigma$ the sigmoid logistic function. 

We then compute a compatibility score between the internal state $\bm{h}_t$ and the sentinel vector through a single-layer neural network; analogously, we compute a compatibility function between $\bm{h}_t$ and the regions in $\bm{r}_t$.
\begin{equation}
    z^c_t = \bm{w}_{h}^T \tanh(\bm{W}_{sg}\bm{s}^c_t + \bm{W}_{g} \bm{h}_t)
\end{equation}
\begin{equation}
    \bm{z}^r_t = \bm{w}_{h}^T \tanh(\bm{W}_{sr}\bm{r}_t + (\bm{W}_{g} \bm{h}_t)\mathbbm{1}^T)
\end{equation}
where $n$ is the number of regions in $\bm{r}_t$, $\mathbbm{1} \in \mathbb{R}^{n}$ is a vector with all elements set to 1, $\bm{w}_h^T$ is a row vector, and all $\bm{W}_*$, $\bm{w}_*$ are learnable parameters. Notice that the representation extracted from the internal state is shared between all compatibility scores, as if the region set and the sentinel vector were part of the same attentive distribution. Contrarily to an attentive mechanism, however, there is no value extraction.

The probability of shifting from one chunk to the next one is defined as the probability of attending the sentinel vector $\bm{s}^c_t$ in a distribution over $\bm{s}^c_t$ and the regions in $\bm{r}_t$:
\begin{equation}
    p(g_t = 1 | \bm{R}) = \frac{\exp{z^c_t}}{\exp{z^c_t} + \sum_{i=1}^n \exp{\bm{z}^r_{ti}}}
\end{equation}
where $\bm{z}^r_{ti}$ indicates the $i$-th element in $\bm{z}^r_{t}$, and we dropped the dependency between $n$ and $t$ for clarity. At test time, the value of gate $g_t \in \{0, 1\}$ is then sampled from $p(g_t | \bm{R})$ and drives the shifting to the next region set in $\bm{R}$.

\tit{Adaptive attention with visual sentinel}
While the chunk-shifting gate predicts the end of a chunk, thus linking the generation process with the control signal given by $\bm{R}$, once $\bm{r}_t$ has been selected a second mechanism is needed to attend its regions and distinguish between visual and textual words. To this end, we build an adaptive attention mechanism with a visual sentinel~\cite{lu2017knowing}.

The visual sentinel vector models a component of the memory to which the model can fall back when it chooses to not attend a region in $\bm{r}_t$. Analogously to Eq.~\ref{eq:s_g}, it is defined as:
\begin{align}
    \bm{l}^v_t &= \sigma(\bm{W}_{is} \bm{x}_t + \bm{W}_{hs} \bm{h}_{t-1}) \\
    \bm{s}^v_t &= \bm{l}^v_t\odot\tanh(\bm{m}_t)
\end{align}
where $\bm{W}_{is} \in \mathbb{R}^{d\times k}$ and $\bm{W}_{hs} \in \mathbb{R}^{d\times d}$ are matrices of learnable weights. An attentive distribution is then generated over the regions in $\bm{r}_t$ and the visual sentinel vector $\bm{s}^v_t$: 
\begin{equation}
    \bm{\alpha}_t = \text{softmax}([ \bm{z}^r_t; \bm{w}_{h}^T \tanh(\bm{W}_{ss}\bm{s}^v_t + \bm{W}_{g} \bm{h}_t)])
\end{equation}
where $[\cdot]$ indicates concatenation. Based on the attention distribution, we obtain a context vector which can be fed to the LSTM as a representation of what the network is attending:
\begin{equation}
    \bm{c}_t = \sum_{i=1}^{n+1} \bm{\alpha}_{ti} [\bm{r}_t; \bm{s}_t^v]
\end{equation}
Notice that the context vector will be, mostly, an approximation of one of the regions in $\bm{r}_t$ or the visual sentinel. However, $\bm{r}_t$ will vary at different timestep according to the chunk-shifting mechanism, thus following the control input. The model can alternate the generation of visual and textual words by means of the visual sentinel.


\subsection{Objective}
The captioning model is trained using a loss function which considers the two output distributions of the model. Given the target ground-truth caption $\bm{y}^*_{1:T}$, the ground-truth region sets $\bm{r}^*_{1:T}$ and chunk-shifting gate values corresponding to each timestep $g^*_{1:T}$, we train both distributions by means of a cross-entropy loss. The relationship between target region sets and gate values will be further expanded in the implementation details. The loss function for a sample is defined as:
\label{sec:objective}
\begin{align}
    L(\theta) &= -\sum_{t=1}^T \Big( \log  \overbrace{p(y_t^* | \bm{r}^*_{1:t}, \bm{y}^*_{1:t-1})}^{\text{Word-level probability}} +  \nonumber \\ 
    &+ g_t^* \log p(g_t=1 | \bm{r}^*_{1:t}, \bm{y}^*_{1:t-1}) + \nonumber\\
    &+ (1-g_t^*)\log (1-\underbrace{p(g_t=1 | \bm{r}^*_{1:t}, \bm{y}^*_{1:t-1})}_{\text{Chunk-level probability}})\Big)
\end{align}

Following previous works~\cite{ranzato2015sequence,rennie2017self,anderson2018bottom}, after a pre-training step using cross-entropy, we further optimize the sequence generation using Reinforcement Learning. Specifically, we use the self-critical sequence training approach~\cite{rennie2017self}, which baselines the REINFORCE algorithm with the reward obtained under the inference model at test time.

Given the nature of our model, we extend the approach to work on multiple output distributions. At each timestep, we sample from both $p(y_t|\bm{R})$ and $p(g_t|\bm{R})$ to obtain the next word $w_{t+1}$ and region set $\bm{r}_{t+1}$. Once a $\text{EOS}$ tag is reached, we compute the reward of the sampled sentence $\bm{w}^s$ and backpropagate with respect to both the sampled word sequence $\bm{w}^s$ and the sequence of chunk-shifting gates $\bm{g}^s$. The final gradient expression is thus:
\begin{equation}
    \nabla_\theta L(\theta) = -(r(\bm{w}^s)-b) (\nabla_\theta \log p(\bm{w}^s) + \nabla_\theta \log p(\bm{g}^s))
\end{equation}
where $b = r(\hat{\bm{w}})$ is the reward of the sentence obtained using the inference procedure (\ie~by sampling the word and gate value with maximum probability). We then build a reward function which jointly considers the quality of the caption and its alignment with the control signal $\bm{R}$.

\tinytit{Rewarding caption quality}
To reward the overall quality of the generated caption, we use image captioning metrics as a reward. Following previous works~\cite{anderson2018bottom}, we employ the CIDEr metric (specifically, the CIDEr-D score) which has been shown to correlate better with human judgment~\cite{vedantam2015cider}.

\begin{figure*}[t]
\centering
\begin{tabular}{cc}
\includegraphics[width=0.49\textwidth]{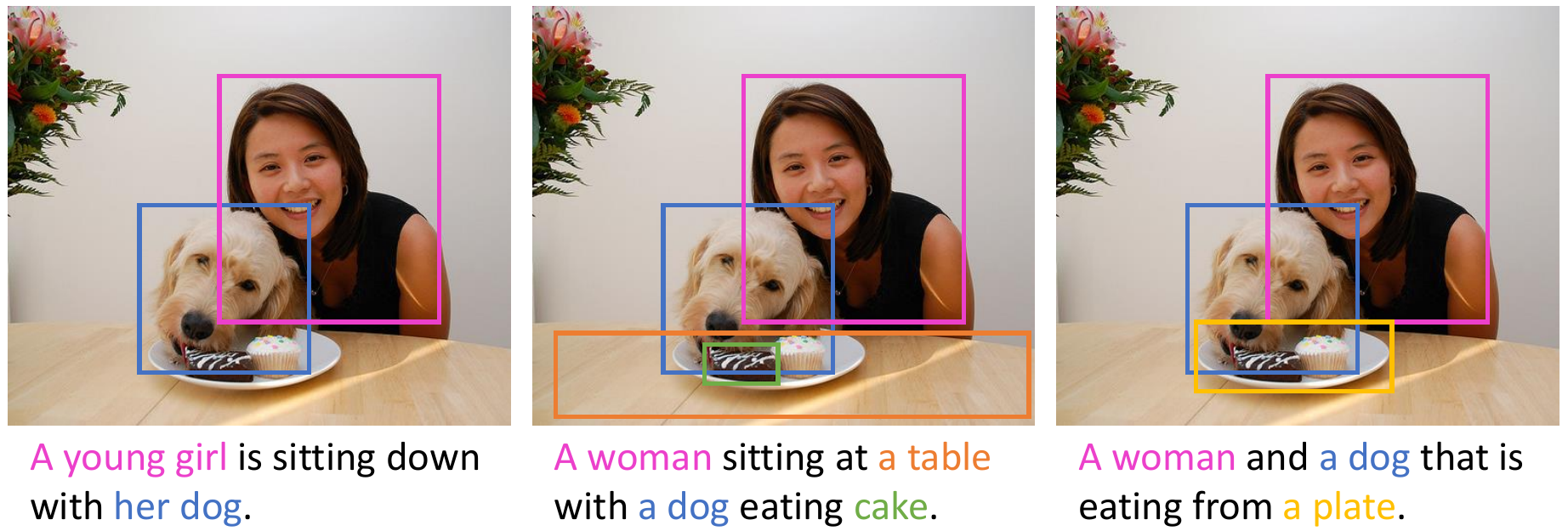} 
\includegraphics[width=0.49\textwidth]{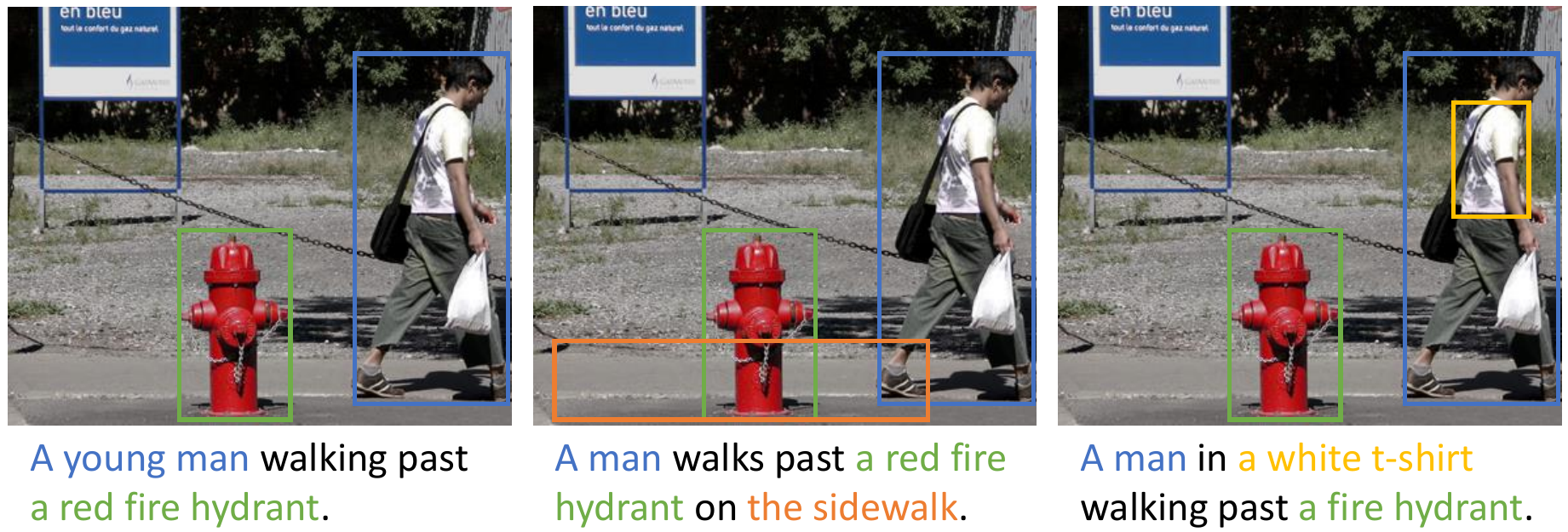}\\ 
\end{tabular}
\caption{\small{Sample captions and corresponding visual groundings from the COCO Entities dataset. Different colors show a correspondence between visual chunks and image regions.}}
\label{fig:coco_entities}
\vspace{-.3cm}
\end{figure*}

\tinytit{Rewarding the alignment}
While captioning metrics can reward the semantic quality of the sentence, none of them can evaluate the alignment with respect to the control input\footnote{Although METEOR creates an alignment with respect to the reference caption, this is done for each unigram, thus mixing semantic and alignment errors.}. Therefore, we introduce an alignment score based on the Needleman-Wunsch algorithm~\cite{needleman1970general}.

Given a predicted caption $\bm{y}$ and its target counterpart $\bm{y}^*$, we extract all nouns from both sentences, and evaluate the alignment between them, recalling the relationships between noun chunks and region sets. We use the following scoring system: the reward for matching two nouns is equal to the cosine similarity between their word embeddings; a gap gets a negative reward equal to the minimum similarity value, \ie~$-1$. Once the optimal alignment is computed, we normalize its score, $al(\bm{y}, \bm{y}^*)$ with respect to the length of the sequences. The alignment score is thus defined as:
\begin{equation}
    \text{NW}(\bm{y}, \bm{y}^*) = \frac{al(\bm{y}, \bm{y}^*)}{\max(\#\bm{y}, \#\bm{y}^*)}
\end{equation}
where $\#\bm{y}$ and $\#\bm{y}^*$ represent the number of nouns contained in $\bm{y}$ and $\bm{y}^*$, respectively. Notice that $\text{NW}(\cdot, \cdot) \in [ -1, 1 ]$. The final reward that we employ is a weighted version of CIDEr-D and the alignment score.


\subsection{Controllability through a set of detections}
\label{sec:sorting_network}
The proposed architecture, so far, can generate a caption controlled by a sequence of region sets $\bm{R}$. To deal with the case in which the control signal is unsorted, \ie~a set of regions sets, we build a \textit{sorting network} which can arrange the control signal in a candidate order, learning from data. The resulting sequence can then be given to the captioning network to produce the output caption (Fig.~\ref{fig:model}).

To this aim, we train a network which can learn a permutation, taking inspiration from Sinkhorn networks~\cite{mena2018learning}. As shown in~\cite{mena2018learning}, the non-differentiable parameterization of a permutation can be approximated in terms of a differentiable relaxation, the so-called Sinkhorn operator. While a permutation matrix has exactly one entry of 1 in each row and each column, the Sinkhorn operator iteratively normalizes rows and columns of any matrix to obtain a ``soft'' permutation matrix, \ie~a real-valued matrix close to a permutation one.

Given a set of region sets $\mathcal{R} = \{ \bm{r}_1, \bm{r}_2, ..., \bm{r}_N \}$, we learn a mapping from $\mathcal{R}$ to its sorted version $\bm{R}^*$. Firstly, we pass each element in $\mathcal{R}$ through a fully-connected network which processes every item of a region set independently and produces a single output feature vector with length $N$. By concatenating together the feature vectors obtained for all region sets, we thus get a $N \times N$ matrix, which is then passed to the Sinkhorn operator to obtain the soft permutation matrix $\bm{P}$. The network is then trained by minimizing the mean square error between the scrambled input and its reconstructed version obtained by applying the soft permutation matrix to the sorted ground-truth, \ie~$\bm{P}^T \bm{R}^*$.

At test time, we take the soft permutation matrix and apply the Hungarian algorithm~\cite{kuhn1955hungarian} to obtain the final permutation matrix, which is then used to get the sorted version of $\mathcal{R}$ for the captioning network.

\subsection{Implementation details}
\tinytit{Language model and image features}
We use a language model with two LSTM layers (Fig.~\ref{fig:model}): the input of the bottom layer is the concatenation of the embedding of the current word, the image descriptor, as well as the hidden state of the second layer. This layer predicts the context vector via the visual sentinel as well as the chunk-gate. The second layer, instead, takes as input the context vector and the hidden state of the bottom layer and predicts the next word. 

To represent image regions, we use Faster R-CNN~\cite{ren2015faster} with ResNet-101~\cite{he2016deep}. In particular, we employ the model finetuned on the Visual Genome dataset~\cite{krishnavisualgenome} provided by~\cite{anderson2018bottom}. As image descriptor, following the same work~\cite{anderson2018bottom}, we average the feature vectors of all the detections.

The hidden size of the LSTM layers is set to $1000$, and that of attention layers to $512$, while the input word embedding size is set to $1000$.

\tinytit{Ground-truth chunk-shifting gate sequences}
Given a sentence where each word of a noun chunk is associated to a region set, we build the chunk-shifting gate sequence $\{ g^*_t\}_t$ by setting $g^*_t$ to $1$ on the last word of every noun chunk, and $0$ otherwise. The region set sequence $\{\bm{r}_t^*\}_t$ is built accordingly, by replicating the same region set until the end of a noun chunk, and then using the region set of the next chunk.
To compute the alignment score and for extracting dependencies, we use the spaCy NLP toolkit\footnote{\url{https://spacy.io/}}. We use GloVe~\cite{pennington2014glove} as word vectors.

\tinytit{Sorting network}
To represent regions, we use Faster R-CNN vectors, the normalized position and size and the GloVe embedding of the region class. Additional details on architectures and training can be found in the Supplementary material.


\begin{table}[t!]
\small
\centering
\setlength{\tabcolsep}{.25em}
\begin{tabular}{lccc}
\toprule 
\textbf{COCO Entities (ours)} & Train & Validation & Test \\
\midrule
Nb. of captions                     & 545,202 & 7,818 & 7,797 \\
Nb. of noun chunks                  & 1,518,667 & 20,787 & 20,596 \\
Nb. of noun chunks per caption      & 2.79 & 2.66 & 2.64 \\
Nb. of unique classes               & 1,330 & 725 & 730 \\
\toprule
\textbf{Flickr30k Entities} & Train & Validation & Test \\
\midrule
Nb. of captions                     & 144,256 & 5,053 & 4,982 \\
Nb. of noun chunks                  & 416,018 & 14,626 & 14,556 \\
Nb. of noun chunks per caption      & 2.88 & 2.89 & 2.92 \\
Nb. of unique classes               & 1,026 & 465 & 458 \\
\toprule
\end{tabular}
\caption{Statistics on our COCO Entities dataset, in comparison with those of Flick30k Entities.}
\label{tab:coco_stas}
\vspace{-.3cm}
\end{table}

\section{Experiments}
\label{sec:experiments}

\begin{table*}
\small
\centering
\setlength{\tabcolsep}{.23em}
\resizebox{\linewidth}{!}{
\begin{tabular}{lcccccccccccccccccccc}
\toprule
& \multicolumn{6}{c}{Cross-Entropy Loss} & & \multicolumn{6}{c}{CIDEr Optimization} & & \multicolumn{6}{c}{CIDEr + NW Optimization} \\
\cmidrule{2-7}
\cmidrule{9-14}
\cmidrule{16-21}
Method & B-4 & M & R & C & S & NW & \:\:& B-4 & M & R & C & S & NW & \:\:& B-4 & M & R & C & S & NW \\
\midrule
FC-2K$^\dagger$~\cite{rennie2017self}             & 10.4 & 17.3 & 36.8 & 98.3 & 25.2 & 0.257 & & 12.3 & 18.5 & 39.6 & 117.5 & 26.9 & 0.273 & & - & - & - & - & - & - \\
Up-Down$^\dagger$~\cite{anderson2018bottom}       & 12.9 & 19.3 & 40.0 & 119.9 & 29.3 & 0.296 & & 14.2 & 20.0 & 42.1 & 133.9 & 30.0 & 0.310 & & - & - & - & - & - & - \\
Neural Baby Talk$^\dagger$~\cite{lu2018neural}    & 12.9 & 19.2 & 40.4 & 120.2 & 29.5 & 0.305 & & - & - & - & - & - & - & & - & - & - & - & - & - \\
\midrule
Controllable LSTM                                 & 11.4 & 18.1 & 38.5 & 106.8 & 27.6 & 0.275 & & 12.8 & 18.9 & 40.9 & 123.0 & 28.5 & 0.290 & & 12.9 & 19.3 & 41.3 & 124.0 & 28.9 & 0.341 \\
Controllable Up-Down                              & 17.3 & 23.0 & 46.7 & 161.0 & 39.1 & 0.396 & & 17.4 & 22.9 & 47.1 & 168.5 & 39.0 & 0.397 & & 17.9 & 23.6 & 48.2 & 171.3 & 40.7 & 0.443 \\
\midrule

\ours \textit{w/} single sentinel               & 20.0 & 23.9 & 51.1 & 183.3 & 43.9 & 0.480 & & 21.7 & 25.3 & 54.5 & 202.6 & 47.6 & 0.606 & & 21.3 & 25.3 & 54.5 & 201.1 & 48.1 & 0.648 \\
\ours \textit{w/o} visual sentinel              & 20.8 & \textbf{24.4} & 52.4 & 191.2 & 45.1 & \textbf{0.508} & & 22.2 & 25.4 & 55.0 & 206.2 & 47.6 & 0.607 & & 21.5 & 25.1 & 54.7 & 202.2 & 48.1 & 0.639 \\
\ours                                           & \textbf{20.9} & \textbf{24.4} & \textbf{52.5} & \textbf{193.0} & \textbf{45.3} & \textbf{0.508} & & \textbf{22.5} & \textbf{25.6} & \textbf{55.1} & \textbf{210.1} & \textbf{48.1} & \textbf{0.615} & & \textbf{22.3} & \textbf{25.6} & \textbf{55.3} & \textbf{209.7} & \textbf{48.5} & \textbf{0.649} \\
\bottomrule
\end{tabular}
}
\caption{Controllability via a sequence of regions, on test portion of COCO Entities. NW refers to the visual chunk alignment measure defined in Sec.~\ref{sec:objective}. The $^\dagger$ marker indicates non-controllable methods.}
\label{tab:order_results}
\end{table*}

\begin{table}
\small
\centering
\setlength{\tabcolsep}{.25em}
\resizebox{\linewidth}{!}{
\begin{tabular}{lcccccc}
\toprule
Method & B-4 & M & R & C & S & NW \\
\midrule
Neural Baby Talk$^\dagger$~\cite{lu2018neural}   & 8.5 & 13.5 & 31.7 & 53.9 & 17.9 & 0.090 \\
\midrule
Controllable LSTM                               & 6.5 & 12.6 & 30.2 & 43.5 & 15.8 & 0.124 \\
Controllable Up-Down                            & 10.4 & 15.2 & 35.2 & 69.5 & 21.7 & 0.190 \\
\midrule
\ours \textit{w/} single sentinel               & 10.7 & 16.1 & 38.1 & 76.5 & 22.8 & 0.260 \\
\ours \textit{w/o} visual sentinel              & 11.1 & 15.5 & 37.2 & 74.7 & 22.4 & 0.244 \\
\ours                                           & \textbf{12.5} & \textbf{16.8} & \textbf{38.9} & \textbf{84.0} & \textbf{23.5} & \textbf{0.263} \\
\bottomrule
\end{tabular}
}
\caption{Controllability via a sequence of regions, on the test portion of Flickr30K Entities.}
\label{tab:order_results_f30k}
\vspace{-.3cm}
\end{table}

\subsection{Datasets}
\label{sec:datasets}
We experiment with two datasets: Flickr30k Entities, which already contains the associations between chunks and image regions, and COCO, which we annotate semi-automatically. Table~\ref{tab:coco_stas} summarizes the datasets we use. 

\tinytit{Flickr30k Entities~\cite{plummer2015flickr30k}}
Based on Flickr30k~\cite{young2014image}, it contains $31,000$ images annotated with five sentences each. Entity mentions in the caption are linked with one or more corresponding bounding boxes in the image. Overall, $276,000$ manually annotated bounding boxes are available. In our experiments, we automatically associate each bounding box with the image region with maximum IoU among those detected by the object detector. We use the splits provided by Karpathy~\etal~\cite{karpathy2015deep}. 

\tinytit{COCO Entities}
Microsoft COCO~\cite{lin2014microsoft} contains more than $120,000$ images, each of them annotated with around five crowd-sourced captions. Here, we again follow the splits defined by~\cite{karpathy2015deep} and automatically associate noun chunks with image regions extracted from the detector~\cite{ren2015faster}. 

We firstly build an index associating each noun of the dataset with the five most similar class names, using word vectors. Then, each noun chunk in a caption is associated by using either its name or the base form of its name, with the first class found in the index which is available in the image. This association process, as confirmed by an extensive manual verification step, is generally reliable and produces few false positive associations. Naturally, it can result in region sets with more than one element (as in Flickr30k), and noun chunks with an empty region set. In this case, we fill empty training region sets with the most probable detections of the image and let the adaptive attention mechanism learn the corresponding association; in validation and testing, we drop those captions. 
Some examples of the additional annotations extracted from COCO are shown in Fig.~\ref{fig:coco_entities}.

\begin{figure*}[t]
\centering
\setlength{\tabcolsep}{.08em}
\def\arraystretch{1.3}
\begin{tabular}[t]{ccc}
\includegraphics[width=0.33\linewidth,valign=t]{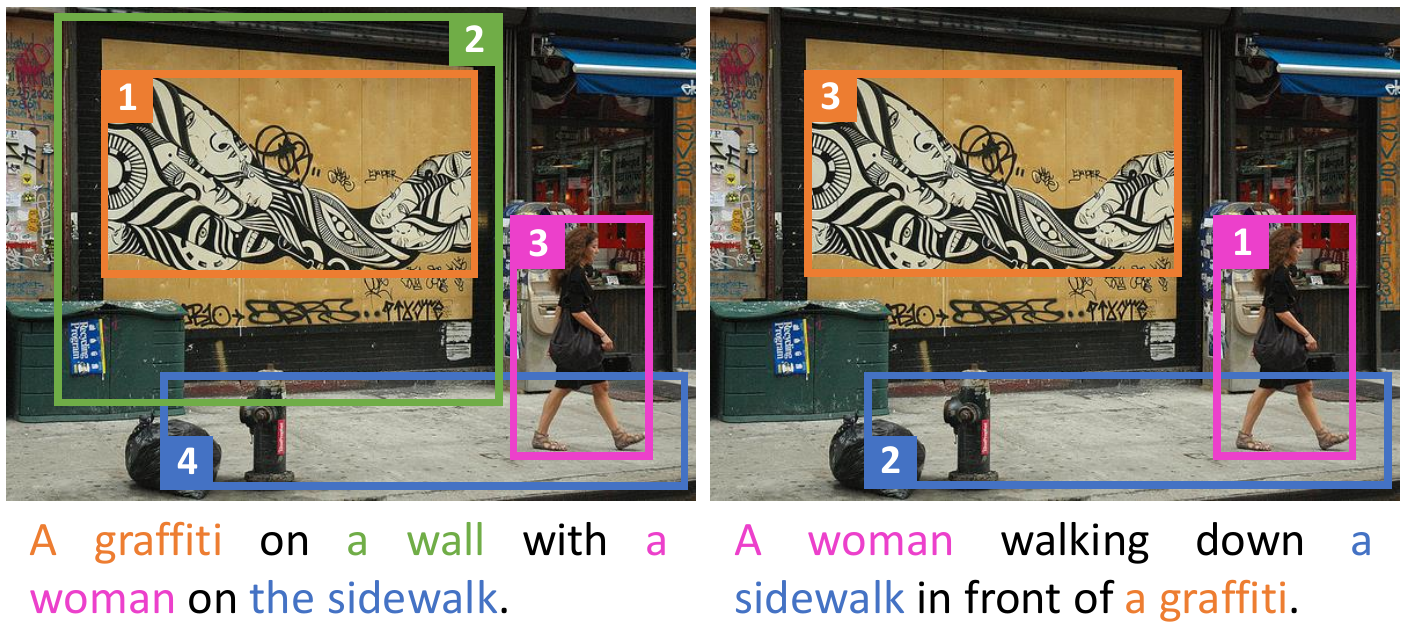} & 
\includegraphics[width=0.33\linewidth,valign=t]{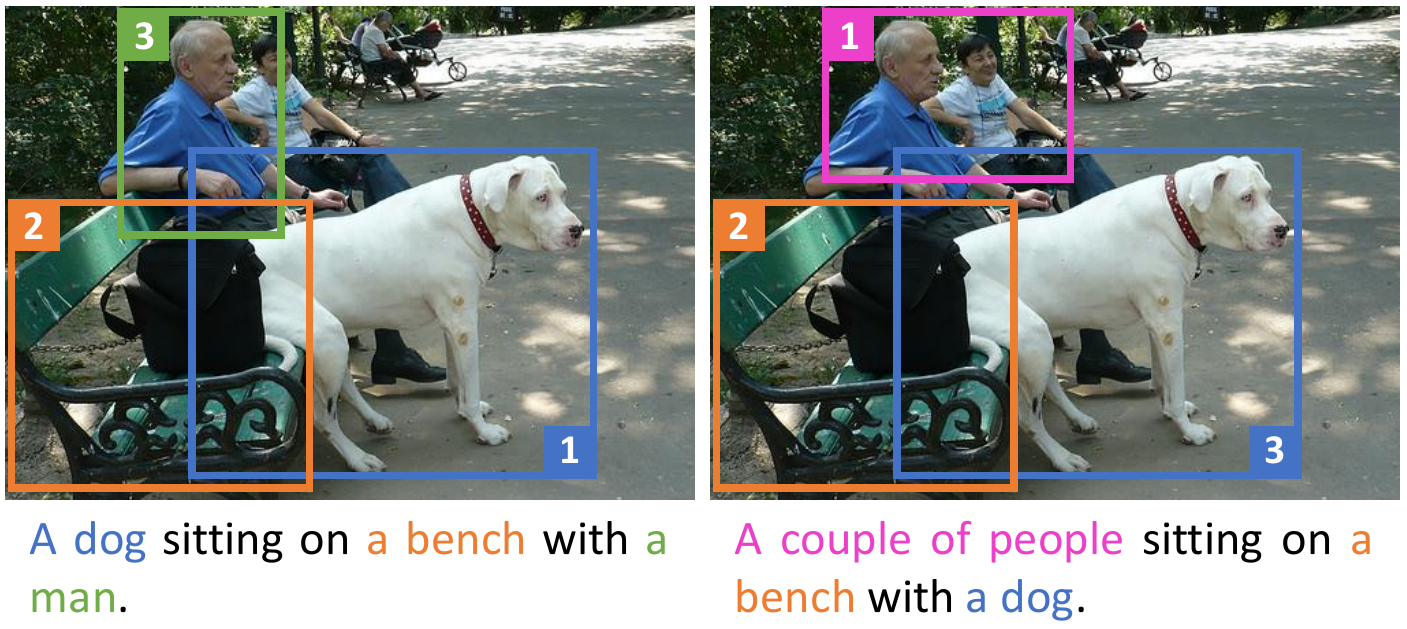} & 
\includegraphics[width=0.33\linewidth,valign=t]{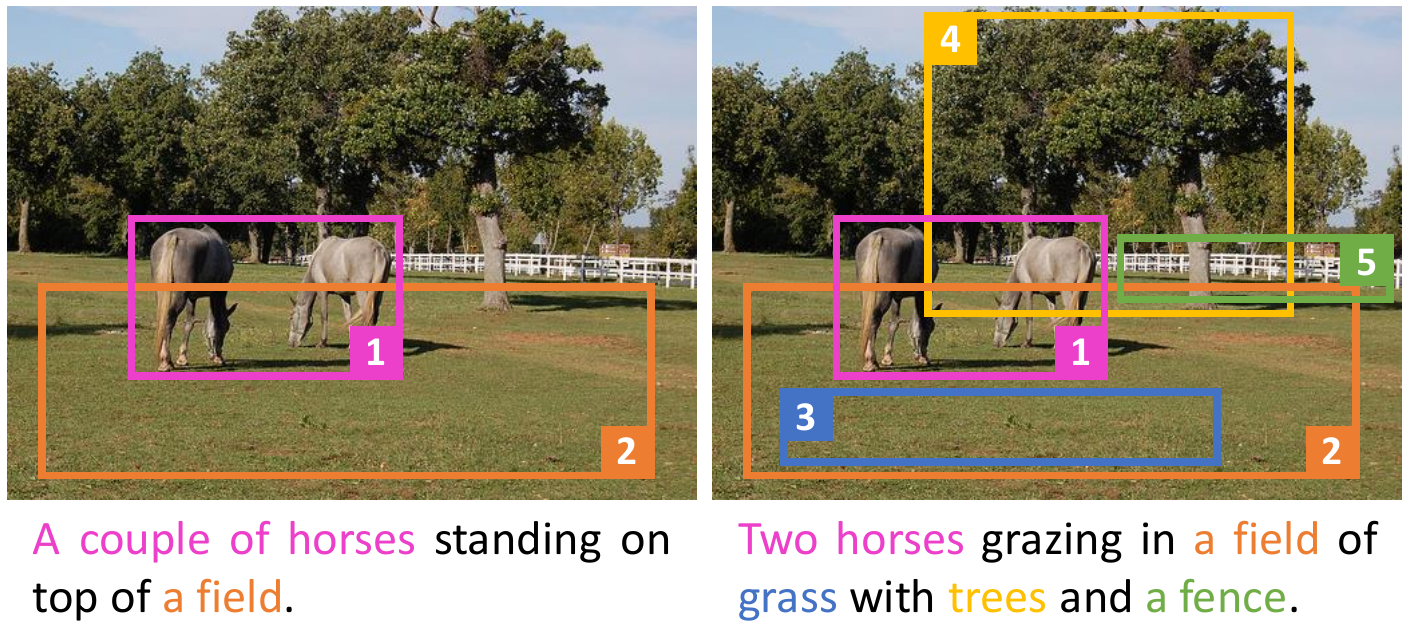}
\end{tabular}
\caption{Sample results of controllability via a sequence of regions. Different colors and numbers show the control sequence and the associations between chunks and regions.}
\label{fig:coco_results1}
\vspace{-.3cm}
\end{figure*}

\subsection{Experimental setting}
The experimental settings we employ is different from that of standard image captioning. In our scenario, indeed, the sequence of set of regions is a second input to the model which shall be consider when selecting the ground-truth sentences to compare against. Also, we employ additional metrics beyond the standard ones like BLEU-4~\cite{papineni2002bleu}, METEOR~\cite{banerjee2005meteor}, $\text{ROUGE}$~\cite{lin2004rouge},  CIDEr~\cite{vedantam2015cider} and SPICE~\cite{spice2016}.

When evaluating the controllability with respect to a sequence, for each ground-truth regions-image input $(\bm{R}, \bm{I})$, we evaluate against all captions in the dataset which share the same pair. Also, we employ the alignment score (NW) to evaluate how the model follows the control input. 

Similarly, when evaluating the controllability with respect to a set of regions, given a set-image pair $(\mathcal{R}, \bm{I})$, we evaluate against all ground-truth captions which have the same input. To assess how the predicted caption covers the control signal, we also define a soft intersection-over-union (IoU) measure between the ground-truth set of nouns and its predicted counterpart, recalling the relationships between region sets and noun chunks. Firstly, we compute the optimal assignment between the two set of nouns, using distances between word vectors and the Hungarian algorithm~\cite{kuhn1955hungarian}, and define an intersection score between the two sets as the sum of assignment profits. Then, recalling that set union can be expressed in function of an intersection, we define the IoU measure as follows:
\begin{equation}
    \text{IoU}(\bm{y}, \bm{y}^*) = \frac{\text{I}(\bm{y}, \bm{y}^*)}{\#\bm{y} + \#\bm{y}^* - \text{I}(\bm{y}, \bm{y}^*)}
\end{equation}
where $\text{I}(\cdot, \cdot)$ is the intersection score, and the $\#$ operator represents the cardinality of the two sets of nouns.

\begin{figure*}[t]
\centering
\setlength{\tabcolsep}{.08em}
\def\arraystretch{1.3}
\begin{tabular}[t]{ccc}
\includegraphics[width=0.33\linewidth,valign=t]{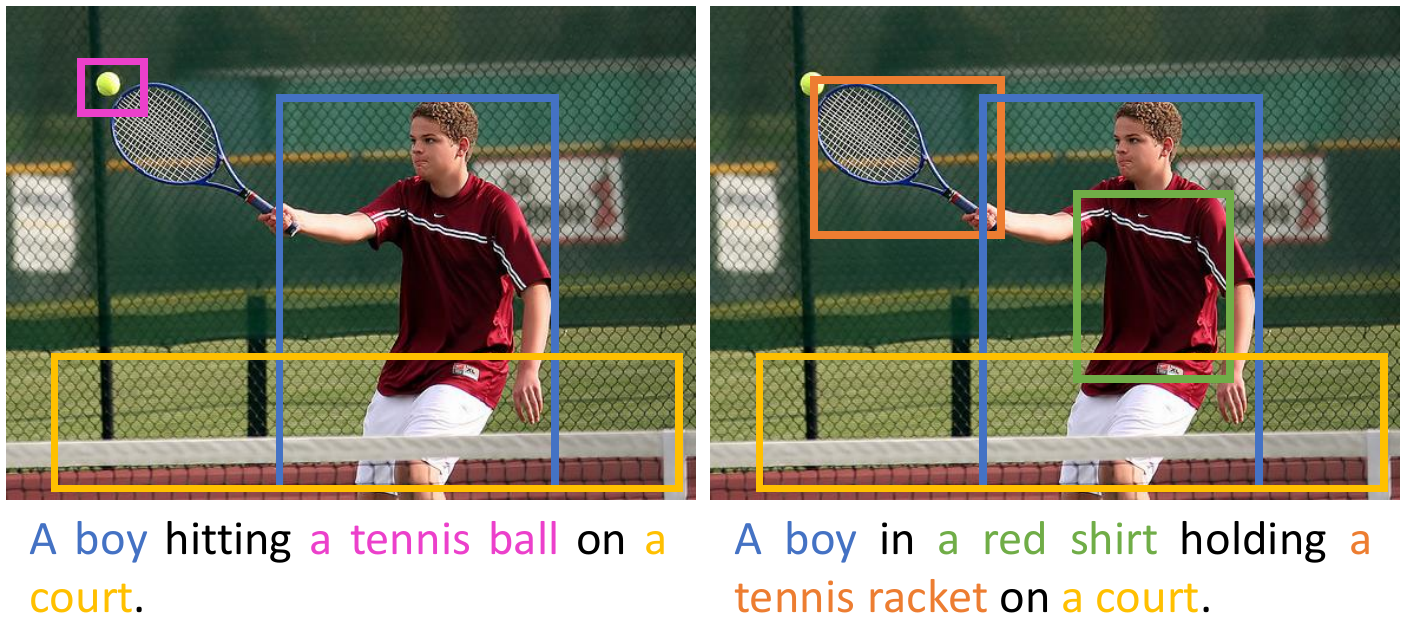} & 
\includegraphics[width=0.33\linewidth,valign=t]{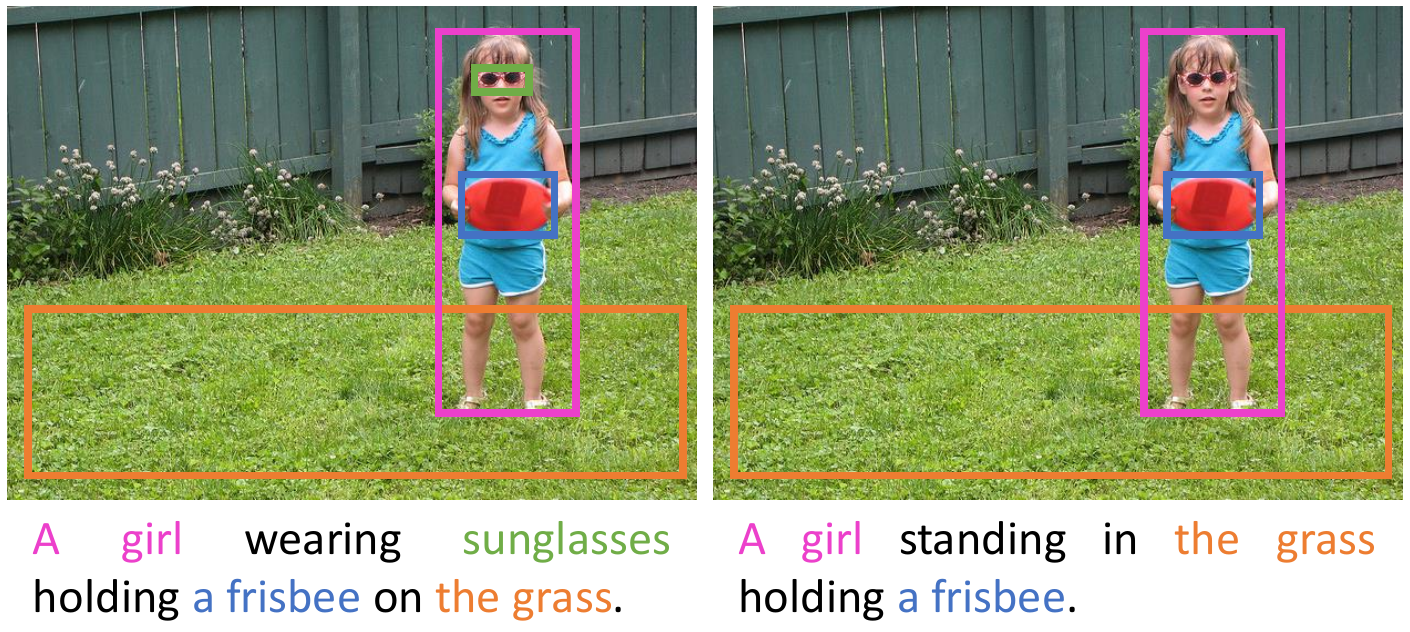} & 
\includegraphics[width=0.33\linewidth,valign=t]{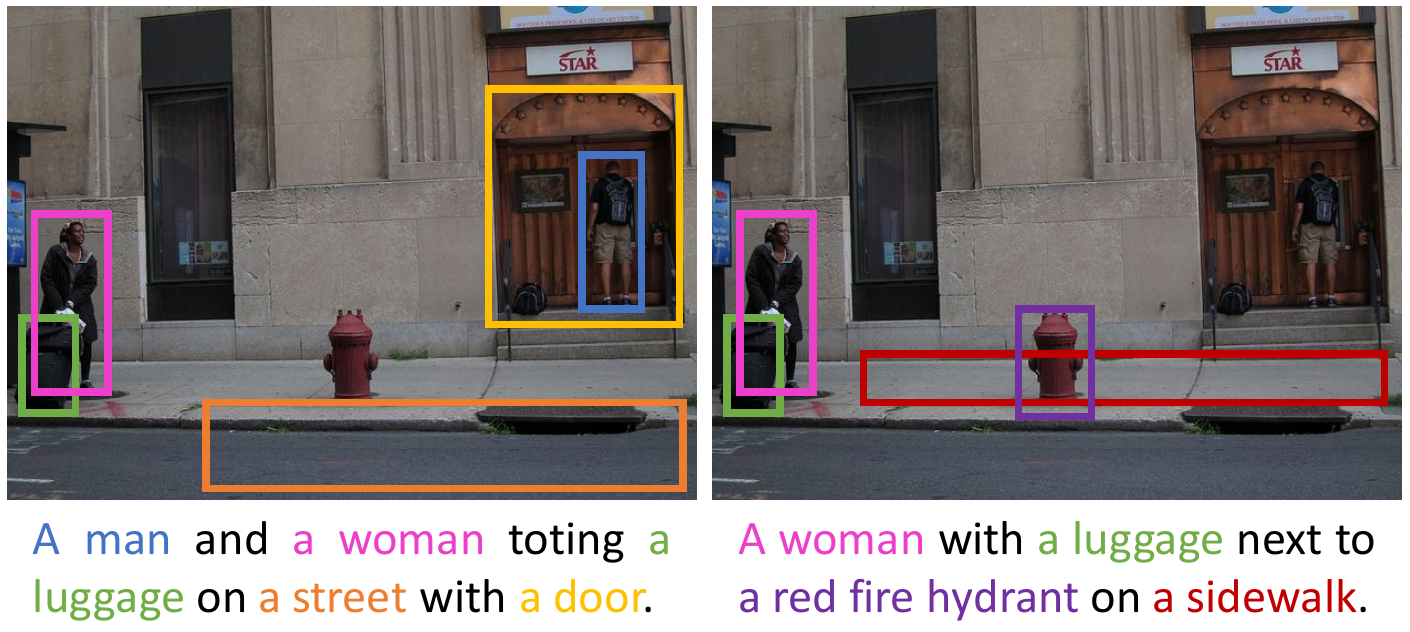}
\end{tabular}
\caption{Sample results of controllability via a set of regions. Different colors show the control set and the associations between chunks and regions.}
\label{fig:coco_results2}
\vspace{-.3cm}
\end{figure*}
 
\subsection{Baselines}
\label{sec:baselines}

\tinytit{Controllable LSTM}
We start from a model without attention: an LSTM language model with a single visual feature vector. Then, we generate a sequential control input by feeding a flattened version of $\bm{R}$ to a second LSTM and taking the last hidden state, which is concatenated to the visual feature vector. The structure of the language model resembles that of~\cite{anderson2018bottom}, without attention.

\tinytit{Controllable Up-Down}
In this case, we employ the full Up-Down model from~\cite{anderson2018bottom}, which creates an attentive distribution over image regions and make it controllable by feeding only the regions selected in $\bm{R}$ and ignoring the rest. This baseline is not sequentially controllable.

\tinytit{\ours without visual sentinel}
To investigate the role of the visual sentinel and its interaction with the gate sentinel, in this baseline we ablate our model by removing the visual sentinel. The resulting baseline, therefore, lacks a mechanism to distinguish between visual and textual words.

\tinytit{\ours with single sentinel} 
Again, we ablate our model by merging the visual and chunk sentinel: a single sentinel is used for both roles, in place of $\bm{s}_t^c$ and $\bm{s}_t^v$. 

\noindent
As further baselines, we also compare against non-controllable captioning approaches, like FC-2K~\cite{rennie2017self}, Up-Down~\cite{anderson2018bottom}, and Neural Baby Talk~\cite{lu2018neural}.

\subsection{Quantitative results}
\tinytit{Controllability through a sequence of detections}
Firstly, we show the performance of our model when providing the full control signal as a sequence of region sets.
Table~\ref{tab:order_results} shows results on COCO Entities, in comparison with the aforementioned approaches. We can see that our method achieves state of the art results on all automatic evaluation metrics, outperforming all baselines both in terms of overall caption quality and in terms of alignment with the control signal. Using the cross-entropy pre-training, we outperform the Controllable LSTM and Controllable Up-Down by 32.0 on CIDEr and 0.112 on NW. Optimizing the model with CIDEr and NW further increases the alignment quality while maintaining outperforming results on all metrics, leading to a final 0.649 on NW, which outperforms the Controllable Up-Down baseline by a 0.25. Recalling that NW ranges from $-1$ to $1$, this improvement amounts to a $12.5\%$ of the full metric range.

In Table~\ref{tab:order_results_f30k}, we instead show the results of the same experiments on Flickr30k Entities, using CIDEr+NW optimization for all controllable methods. Also on this manually annotated dataset, our method outperforms all the compared approaches by a significant margin, both in terms of caption quality and alignment with the control signal.

\begin{table}[t]
\small
\centering
\setlength{\tabcolsep}{.25em}
\resizebox{\linewidth}{!}{
\begin{tabular}{lcccccc}
\toprule
Method & B-4 & M & R & C & S & IoU \\
\midrule
FC-2K$^\dagger$~\cite{rennie2017self}           & 12.5 & 18.5 & 39.6 & 116.5 & 26.6 & 61.0 \\
Up-Down$^\dagger$~\cite{anderson2018bottom}     & 14.4 & 20.0 & 42.2 & 132.8 & 29.7 & 63.2 \\
Neural Baby Talk$^\dagger$~\cite{lu2018neural}  & 13.1 & 19.2 & 40.5 & 119.1 & 29.2 & 62.6 \\
\midrule
Controllable LSTM                               & 12.9 & 19.3 & 41.3 & 123.4 & 28.7 & 64.2 \\
Controllable Up-Down                            & \textbf{18.1} & 23.6 & 48.4 & 170.5 & 40.4 & 71.6 \\
\midrule
\ours \textit{w/} single sentinel               & 17.4 & 23.6 & 48.4 & 168.4 & 43.7 & 75.4 \\
\ours \textit{w/o} visual sentinel              & 17.6 & 23.4 & 48.5 & 168.9 & 43.6 & 75.3 \\
\ours                                           & 18.0 & \textbf{23.8} & \textbf{48.9} & \textbf{173.3} & \textbf{44.1} & \textbf{75.5} \\
\bottomrule
\end{tabular}
}
\caption{Controllability via a set of regions, on the test portion of COCO Entities.}
\label{tab:coco_set_results}
\vspace{-.3cm}
\end{table}

\tinytit{Controllability through a set of detections}
We then assess the performance of our model when controlled with a set of detections.
Tables~\ref{tab:coco_set_results} and \ref{tab:flickr_set_results} show the performance of our method in this setting, respectively on COCO Entities and Flickr30k Entities. We notice that the proposed approach outperforms all baselines and compared approaches in terms of IoU, thus testifying that we are capable of respecting the control signal more effectively. This is also combined with better captioning metrics, which indicate higher semantic quality.

\begin{table}[t]
\small
\centering
\setlength{\tabcolsep}{.25em}
\resizebox{\linewidth}{!}{
\begin{tabular}{lcccccc}
\toprule
Method & B-4 & M & R & C & S & IoU \\
\midrule
Neural Baby Talk$^\dagger$~\cite{lu2018neural}  & 8.6 & 13.5 & 31.9 & 53.8 & 17.8 & 49.9 \\
\midrule
Controllable LSTM                               & 6.4 & 12.5 & 30.2 & 42.9 & 15.6 & 50.8 \\
Controllable Up-Down                            & 10.5 & 15.2 & 35.5 & 69.5 & 21.6 & 54.8 \\
\midrule
\ours \textit{w/} single sentinel               & 9.5 & 15.2 & 35.8 & 65.6 & 21.2 & \textbf{55.0} \\
\ours \textit{w/o} visual sentinel              & 9.8 & 14.8 & 35.0 & 64.2 & 20.9 & 54.3\\
\ours                                           & \textbf{10.9} & \textbf{15.8} & \textbf{36.2} & \textbf{70.4} & \textbf{21.8} & \textbf{55.0} \\
\bottomrule
\end{tabular}
}
\caption{Controllability via a set of regions, on the test portion of Flickr30K Entities.}
\label{tab:flickr_set_results}
\end{table}

\begin{table}
\small
\centering
\setlength{\tabcolsep}{.25em}
\begin{tabular}{lcccccc}
\toprule 
Method & Samples & B-4 & M & R & C & S \\
\toprule
AG-CVAE~\cite{wang2017diverse}                  & 20 &  \textbf{47.1} & 30.9 & 63.8 & 130.8 & 24.4 \\
POS~\cite{deshpande2018diverse}                 & 20 & 44.9 & 36.5 & 67.8 & 146.8 & 27.7 \\
\midrule
\ours                                           & 20 & 44.8 & \textbf{36.6} & \textbf{68.9} & \textbf{156.5} & \textbf{30.9} \\
\bottomrule
\end{tabular}
\caption{Diversity performance on the test portion of COCO.}
\label{tab:coco_oracle_results}
\vspace{-.3cm}
\end{table}

\tinytit{Diversity evaluation}
Finally, we also assess the diversity of the generated captions, comparing with the most recent approaches that focus on diversity. In particular, the variational autoencoder proposed in~\cite{wang2017diverse} and the approach of~\cite{deshpande2018diverse}, which allows diversity and controllability by feeding PoS sequences. To test our method on a significant number of diverse captions, given an image we take all regions which are found in control region sets, and take the permutations which result in captions with higher log-probability. This approach is fairly similar to the sampling strategy used in~\cite{deshpande2018diverse}, even if ours considers region sets. Then, we follow the experimental approach defined in~\cite{wang2017diverse,deshpande2018diverse}: each ground-truth sentence is evaluated against the generated caption with the maximum score for each metric. Higher scores, thus, indicate that the method is capable of sampling high accuracy captions.
Results are reported in Table~\ref{tab:coco_oracle_results}, where to guarantee the fairness of the comparison, we run this experiments on the full COCO test split. As it can be seen, our method can generate significantly diverse captions.

\section{Conclusion}
\label{sec:conclusion}
We presented \textit{Show, Control and Tell}, a framework for generating controllable and grounded captions through regions. Our work is motivated by the need of bringing captioning systems to more complex scenarios. The approach considers the decomposition of a sentence into noun chunks, and grounds chunks to image regions following a control signal. Experimental results, conducted on Flickr30k and on COCO Entities, validate the effectiveness of our approach in terms of controllability and diversity.

{\small
\bibliographystyle{ieee}
\bibliography{egbib}
}

\newpage

\appendix
\section{Sorting network}
\begin{figure}[t]
\centering
\includegraphics[width=\linewidth]{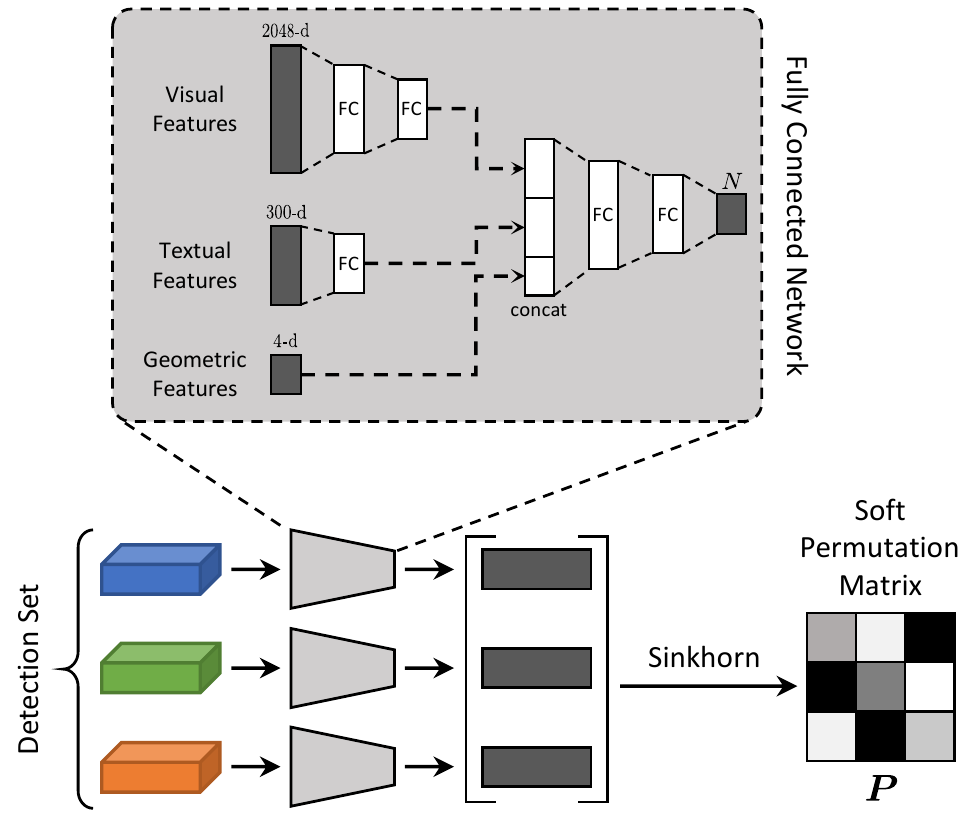} 
\caption{Schema of the sorting network.}
\label{fig:sorting_network}
\vspace{-.2cm}
\end{figure}

We provide additional details on the architecture and training strategy of the sorting network. For the ease of the reader, a schema is reported in Fig.~\ref{fig:sorting_network}.
Given a scrambled sequence of $N$ region sets, each region is encoded through a fully connected network which returns a $N$-dimensional descriptor. The fully connected network employs visual, textual and geometric features: the Faster R-CNN vector of the detection ($2048$-d), the GloVe embedding of the region class ($300$-d) and the normalized position and size of the bounding-box ($4$-d). The visual vector is processed by two layers ($512$-d, $128$-d), while the textual feature is processed by a single layer ($128$-d). The outputs of the visual and textual branches are then concatenated with the geometric features and fed through another fully connected layer ($256$-d). A final layer produces the resulting $N$-dimensional descriptors. All layers have ReLU activations, except for the last fully-connected which has a $\tanh$ activation. In case the region set contains more than one detection, we average-pool the resulting $N$-dimensional descriptors to obtain a single feature vector for a region set.

Once the feature vectors of the scrambled sequence are concatenated, we get a $N \times N$ matrix, which is then converted into a ``soft'' permutation matrix $\bm{P}$ through the Sinkhorn operator. The operator processes a $N$-dimensional square matrix $\bm{X}$ by applying $L$ consecutive row-wise and column-wise normalization, as follows:
\begin{align}
    S^0(\bm{X}) &= \exp(\bm{X}) \\
    S^l(\bm{X}) &= \mathcal{T}c(\mathcal{T}_r(S^{l-1}(\bm{X}))) \\
    \bm{P} &:= S^L(\bm{X})
\end{align}
where $\mathcal{T}_r(\bm{X}) = \bm{X} \oslash (\bm{X} \bm{1}_N \bm{1}_N^T)$, and $\mathcal{T}_c(\bm{X}) = \bm{X} \oslash (\bm{1}_N \bm{1}_N^T \bm{X})$ are the row-wise and column-wise normalization operators, with $\oslash$ denoting element-wise division, $\bm{1}_N$ a column vector of $N$ ones. At test time, once $L$ normalizations ($L=20$ in our experiments) have been performed, the resulting ``soft'' permutation matrix can be converted into a permutation matrix via the Hungarian algorithm~\cite{kuhn1955hungarian}.

At training time, instead, we measure the mean square error between the scrambled sequence and its reconstructed version obtained by applying the soft permutation matrix to the sorted ground-truth sequence $\bm{R}^*$, \ie~$\bm{P}^T \bm{R}^*$. On the implementation side, all tensors are appropriately masked to deal with variable-length sequences and sets. We set the maximum length of input scrambled sequences to $10$. 

\begin{table}[t]
\centering
\small
\setlength{\tabcolsep}{.25em}
\resizebox{\linewidth}{!}{
\begin{tabular}{lccccc}
\toprule
& \multicolumn{2}{c}{COCO Entities} & & \multicolumn{2}{c}{Flickr Entities} \\
\cmidrule{2-3}
\cmidrule{5-6}
& Accuracy & Kendall's Tau & & Accuracy & Kendall's Tau \\
\midrule
Predefined local (det. prob.) & 36.2\% & 0.145 & & 40.0\% & 0.249 \\
Predefined global (det. class)  & 59.1\% & 0.525 & & 58.4\% & 0.565 \\
SVM Rank                        & 54.6\% & 0.448 & & 49.5\% & 0.418 \\
Sinkhorn Network                & \textbf{67.1\%} & \textbf{0.613} & & \textbf{65.2\%} & \textbf{0.633} \\
\bottomrule
\end{tabular}
}
\caption{Sorting network: experimental evaluation.}
\label{tab:sinkhorn_results}
\vspace{-.3cm}
\end{table}

\begin{figure*}[t]
\centering
\setlength{\tabcolsep}{.15em}
\def\arraystretch{1.3}
\begin{tabular}[t]{cc}
\includegraphics[width=0.49\linewidth,valign=t]{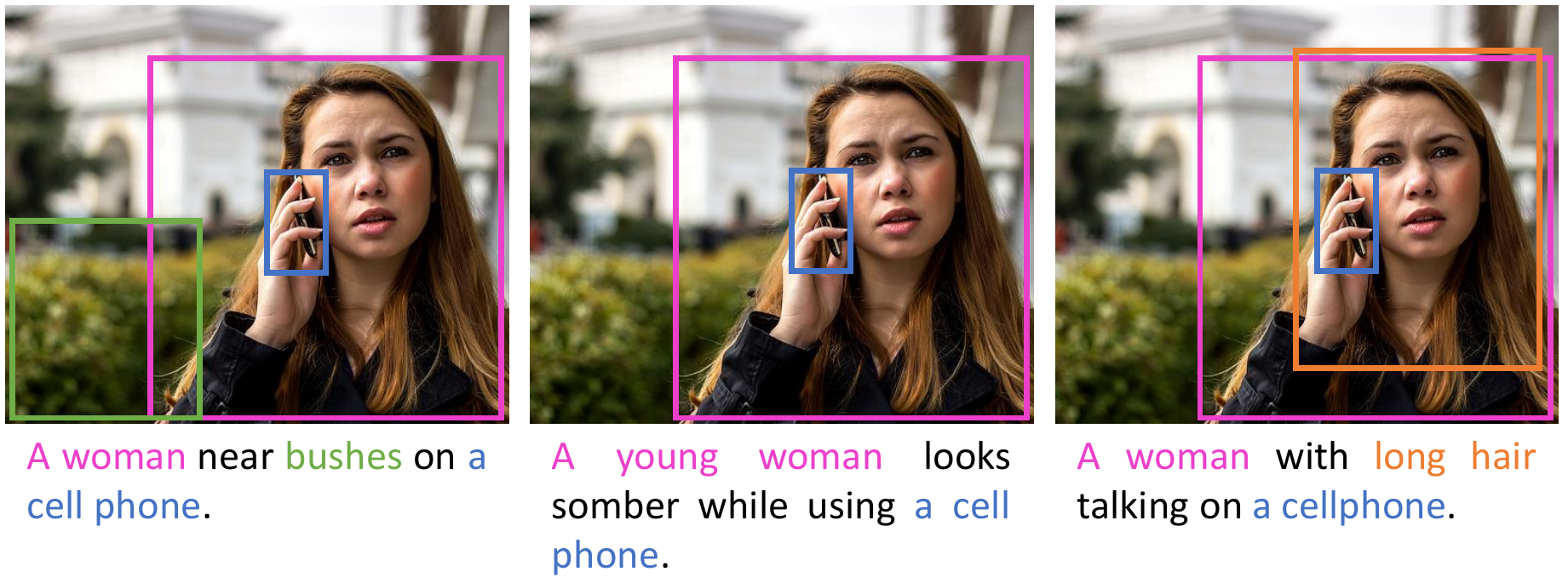} &
\includegraphics[width=0.49\linewidth,valign=t]{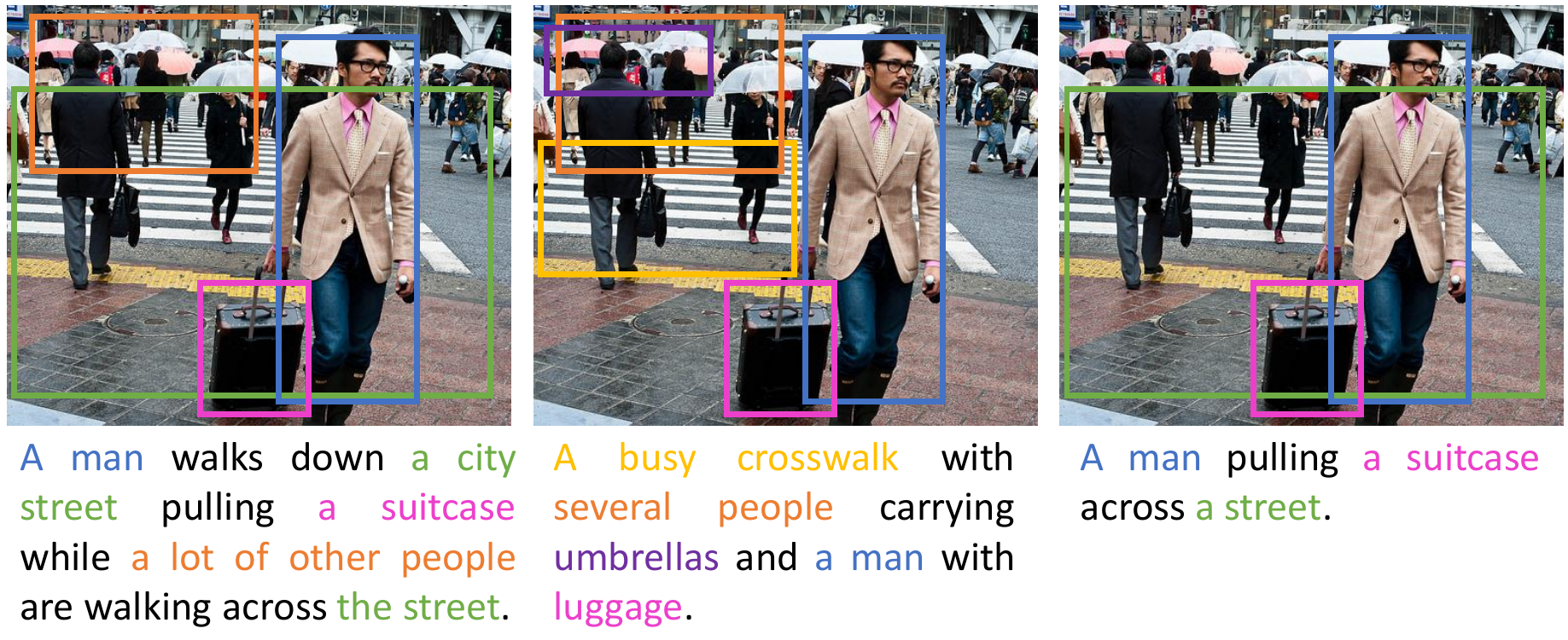} \\
\includegraphics[width=0.49\linewidth,valign=t]{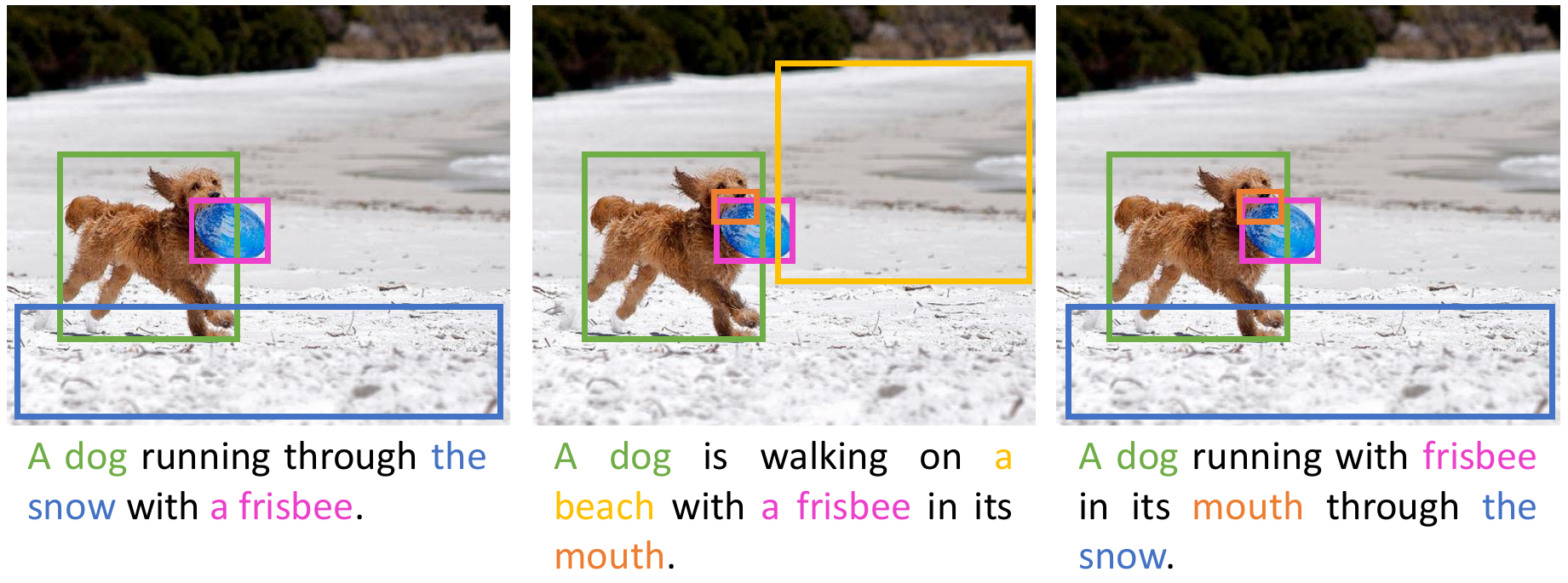} &
\includegraphics[width=0.49\linewidth,valign=t]{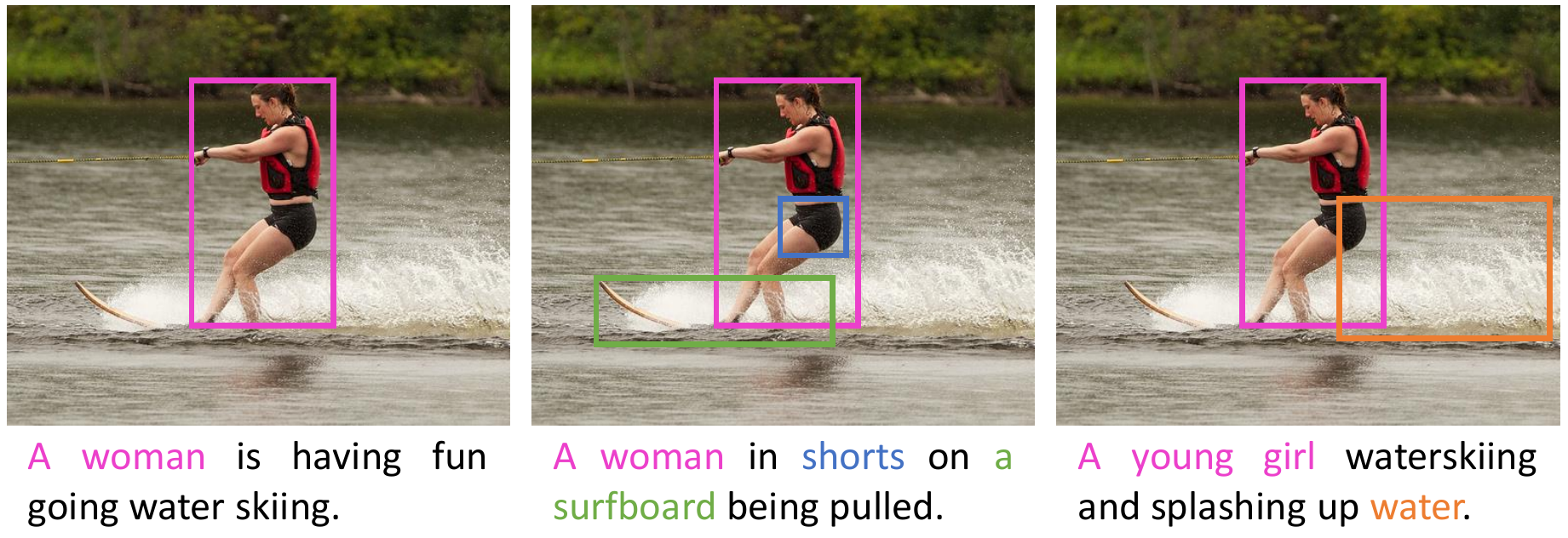} \\
\includegraphics[width=0.49\linewidth,valign=t]{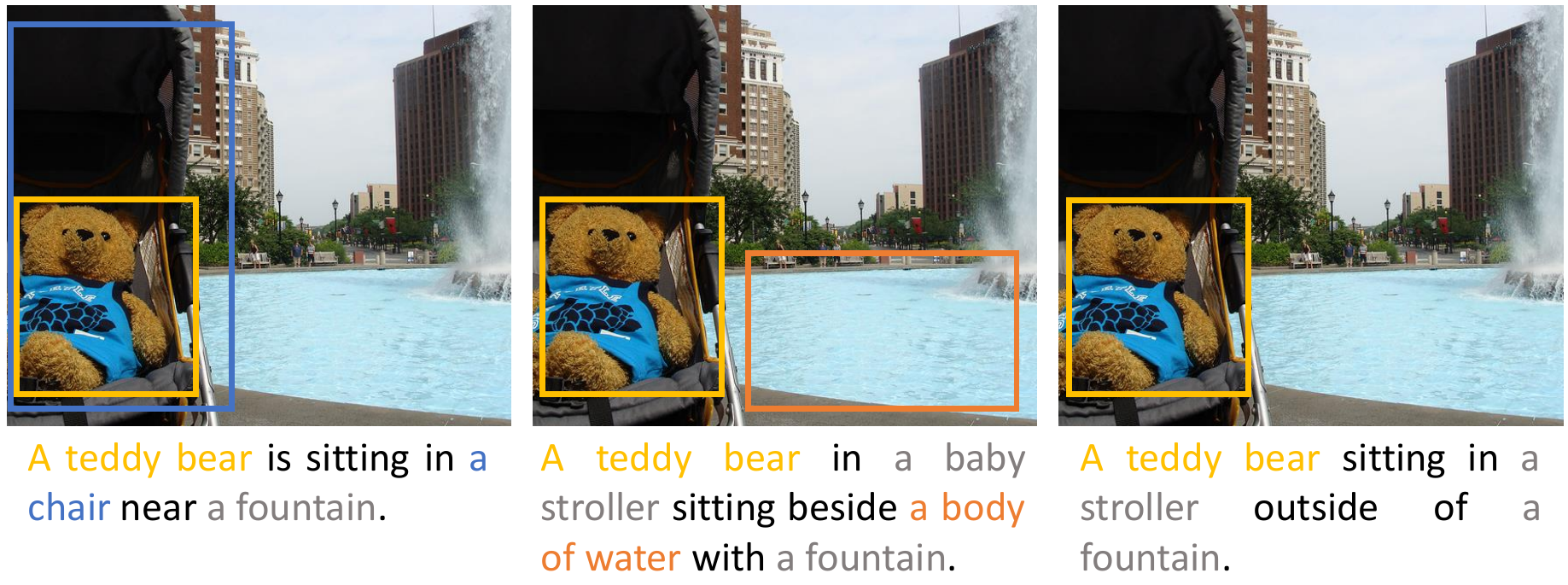} &
\includegraphics[width=0.49\linewidth,valign=t]{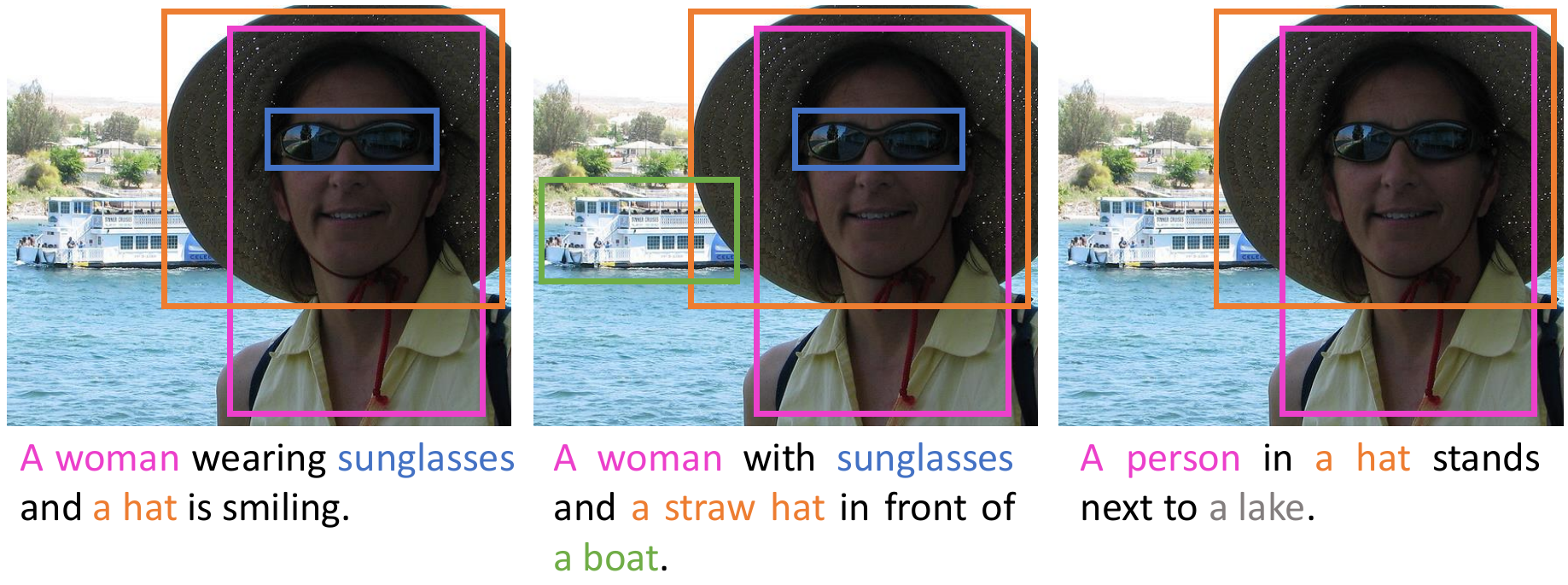} \\
\includegraphics[width=0.49\linewidth,valign=t]{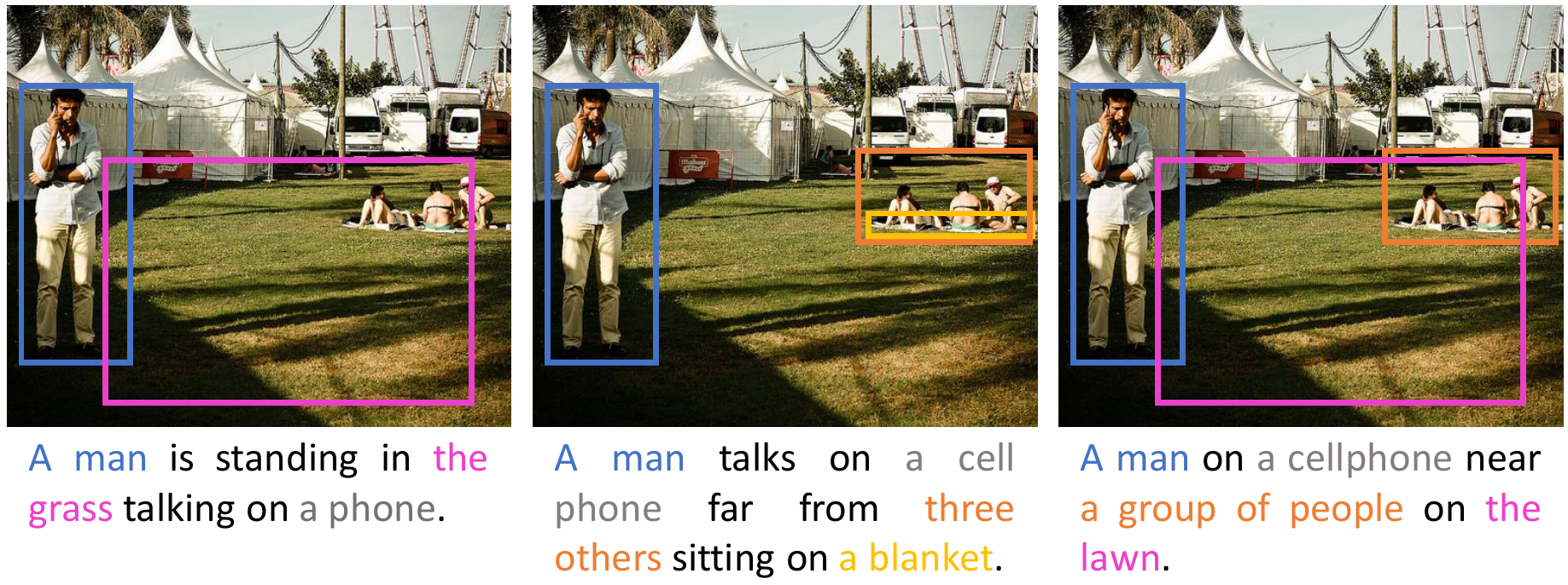} &
\includegraphics[width=0.49\linewidth,valign=t]{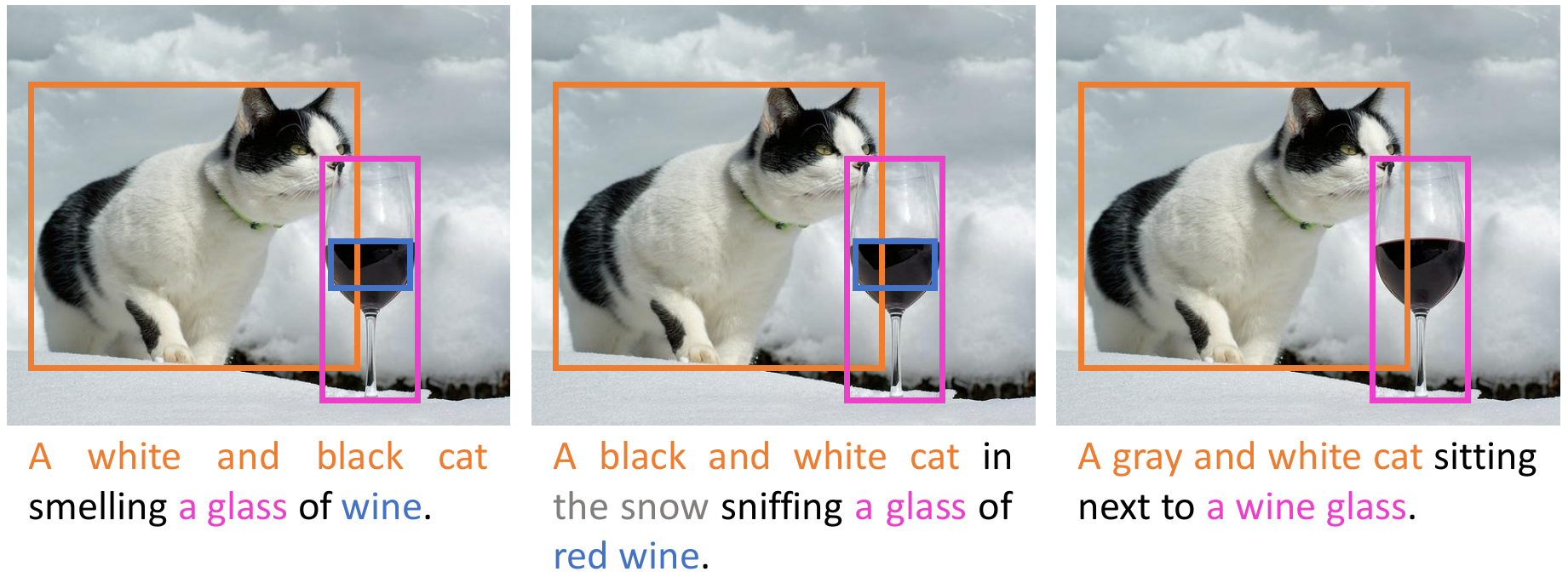} \\
\end{tabular}
\caption{Additional sample captions and corresponding visual groundings from the COCO Entities dataset. Different colors show a correspondence between textual chunks and image regions. Gray color indicates noun chunks for which a visual grounding could not be found, either for missing detections or for errors in the noun-class association.}
\label{fig:coco_entities_supp}
\end{figure*}

In Table~\ref{tab:sinkhorn_results} we evaluate the quality of the rankings in terms of accuracy (proportion of completely correct rankings) and Kendall's Tau (correlation between GT and predicted ranking, between $-1$ and $1$). We compare with a predefined local ranking (sorting detections with their probability), a predefined global ranking based on detection classes, and compare the Sinkorn network with a SVM Rank model trained on the same features. As it can be seen, the Sinkhorn network performs better than other baselines and can generate accurate rankings.

\section{Training details}
We used a weight of 0.2 for the word loss and 0.8 for the two chunk-level terms in Eq. 11. To train both the captioning model and the sorting network, we use the Adam optimizer with an initial learning rate of $5\times10^{-4}$ decreased by a factor of $0.8$ every epoch. For the captioning model, we run the reinforcement learning training with a fixed learning rate of $5\times10^{-5}$. We use a batch size of $100$ for all our experiments. During caption decoding, we employ for all experiments the beam search strategy with a beam size of $5$: similarly to what has been done when training with Reinforcement Learning, we sample from both output distribution to select the most probable sequence of actions. We use early stopping on validation CIDEr for the captioning network, and validation accuracy of the predicted permutations for the sorting network.

\section{The COCO Entities dataset}
In Fig.~\ref{fig:coco_entities_supp}, we report additional examples of the semi-automatic annotation procedure used to collect COCO Entities. As in the main paper, we use different colors to visualize the correspondences between noun chunks and image regions. For the ease of visualization, we display a single region for chunk, even though multiple associations are possible. In this case, the region set would contain more than one element.

In the last two rows, we also report samples in which at least one noun chunk could not be assigned to any detection. Recall that in this case, at training time, we use the most probable detections of the image and let the adaptive attention mechanism learn the corresponding association: we found that this procedure, overall, increases the final accuracy of the network rather than feeding empty region sets. Captions with missing associations are dropped in validation and testing. 

\section{Additional experimental results}
Tables~\ref{tab:order_results_f30k_supp},~\ref{tab:coco_set_results_supp} and~\ref{tab:flickr_set_results_supp} report additional experimental results which have not been reported in the main paper for space constraints.
In particular, Table~\ref{tab:order_results_f30k_supp} integrates Table~3 of the main paper by evaluating the controllability via a sequence of region sets on Flickr30K, when training with cross-entropy only, and when optimizing with CIDEr and CIDEr+NW. Analogously, Tables~\ref{tab:coco_set_results_supp} and~\ref{tab:flickr_set_results_supp} analyze the controllabilty via a set of regions, on both Flickr30K and COCO Entities and with all training strategies.

\begin{table*}
\small
\centering
\setlength{\tabcolsep}{.25em}
\resizebox{\linewidth}{!}{
\begin{tabular}{lcccccccccccccccccccc}
\toprule
& \multicolumn{6}{c}{Cross-Entropy Loss} & & \multicolumn{6}{c}{CIDEr Optimization} & & \multicolumn{6}{c}{CIDEr + NW Optimization} \\
\cmidrule{2-7}
\cmidrule{9-14}
\cmidrule{16-21}
Method & B-4 & M & R & C & S & NW & \:\:& B-4 & M & R & C & S & NW & \:\:& B-4 & M & R & C & S & NW \\
\midrule
Controllable LSTM                               & 6.5 & 12.0 & 29.6 & 40.4 & 15.7 & 0.078 & & 6.7 & 12.1 & 30.0 & 45.5 & 15.8 & 0.079 & & 6.5 & 12.6 & 30.2 & 43.5 & 15.8 & 0.124 \\
Controllable Up-Down                            & 10.1 & 15.2 & 34.9 & 69.2 & 21.6 & \textbf{0.158} & & 10.1 & 14.8 & 35.0 & 69.3 & 21.2 & 0.148 & & 10.4 & 15.2 & 35.2 & 69.5 & 21.7 & 0.190 \\
\midrule
\ours \textit{w/} single sentinel               & 11.0 & \textbf{15.5} & 36.3 & 71.7 & 22.6 & 0.134 & & 11.2 & 15.8 & 37.9 & 77.9 & 22.9 & 0.199 & & 10.7 & 16.1 & 38.1 & 76.5 & 22.8 & 0.260 \\
\ours \textit{w/o} visual sentinel              & 10.8 & 14.9 & 35.4 & 69.3 & 22.2 & 0.142 & & 11.1 & 15.5 & 36.8 & 75.0 & 22.2 & 0.197 & & 11.1 & 15.5 & 37.2 & 74.7 & 22.4 & 0.244 \\
\ours                                           & \textbf{11.3} & 15.4 & \textbf{36.9} & \textbf{74.5} & \textbf{23.4} & 0.152 & & \textbf{12.4} & \textbf{16.6} & \textbf{38.8} & \textbf{83.7} & \textbf{23.5} & \textbf{0.221} & & \textbf{12.5} & \textbf{16.8} & \textbf{38.9} & \textbf{84.0} & \textbf{23.5} & \textbf{0.263} \\
\bottomrule
\end{tabular}
}
\caption{Controllability via a sequence of regions, on the test portion of Flickr30K Entities.}
\label{tab:order_results_f30k_supp}
\end{table*}

\begin{figure*}
\centering
\setlength{\tabcolsep}{.08em}
\def\arraystretch{1.3}
\begin{tabular}[t]{ccc}
\includegraphics[width=0.33\linewidth,valign=t]{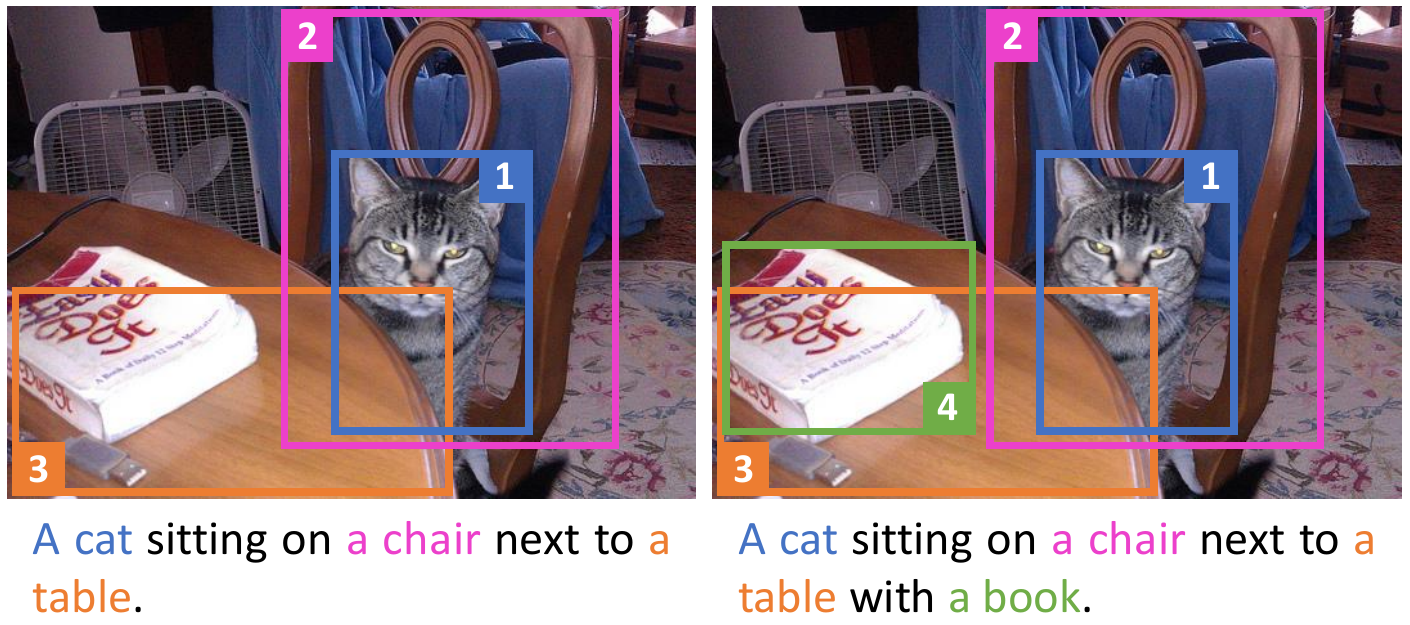} & 
\includegraphics[width=0.33\linewidth,valign=t]{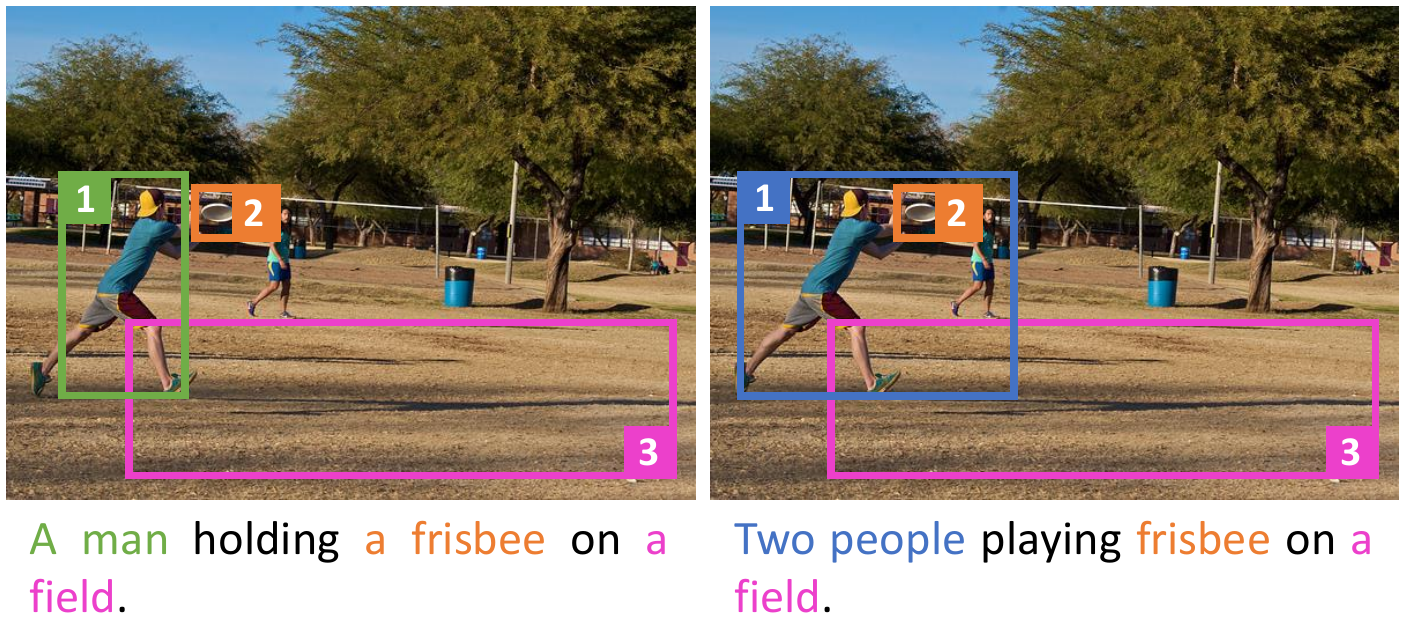} & 
\includegraphics[width=0.33\linewidth,valign=t]{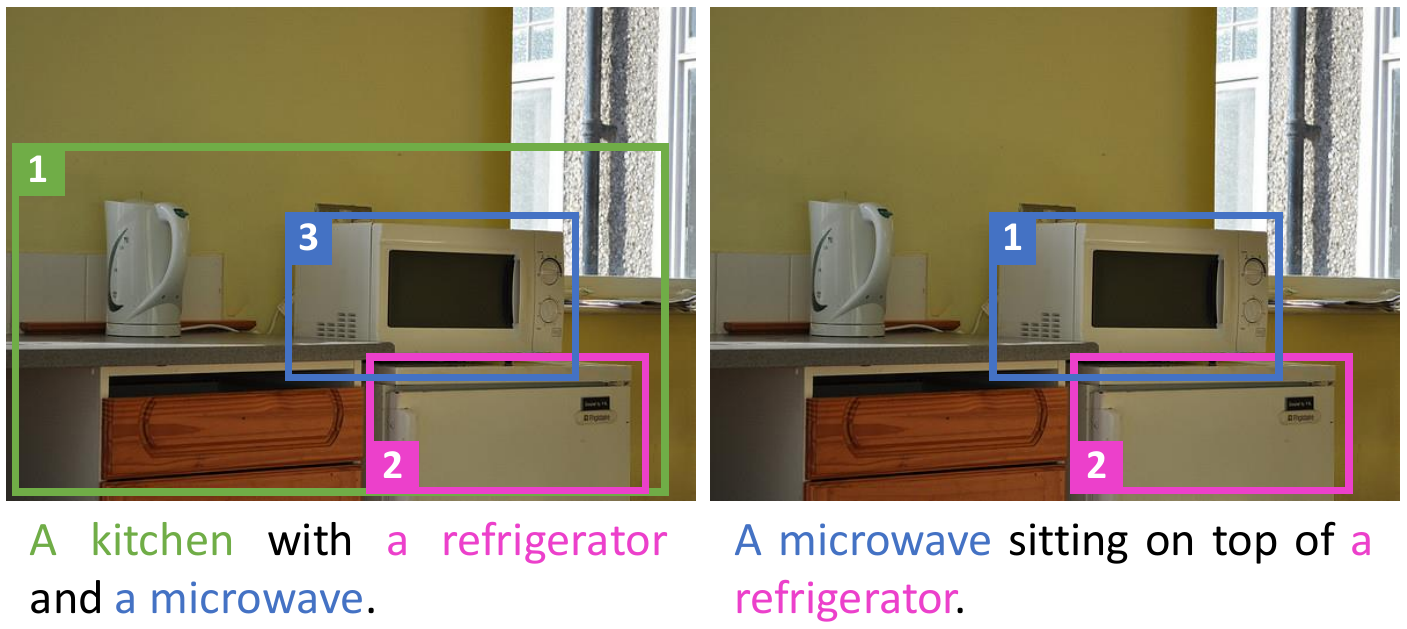} \\
\includegraphics[width=0.33\linewidth,valign=t]{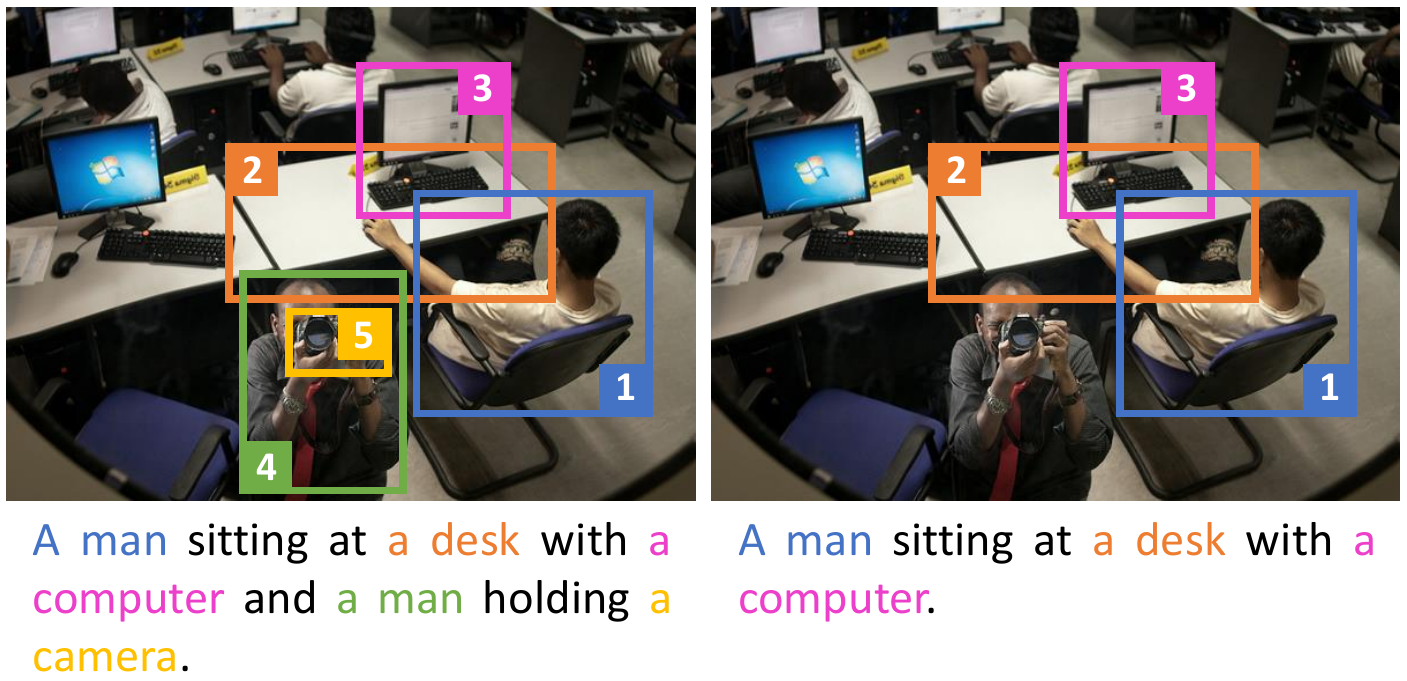} & 
\includegraphics[width=0.33\linewidth,valign=t]{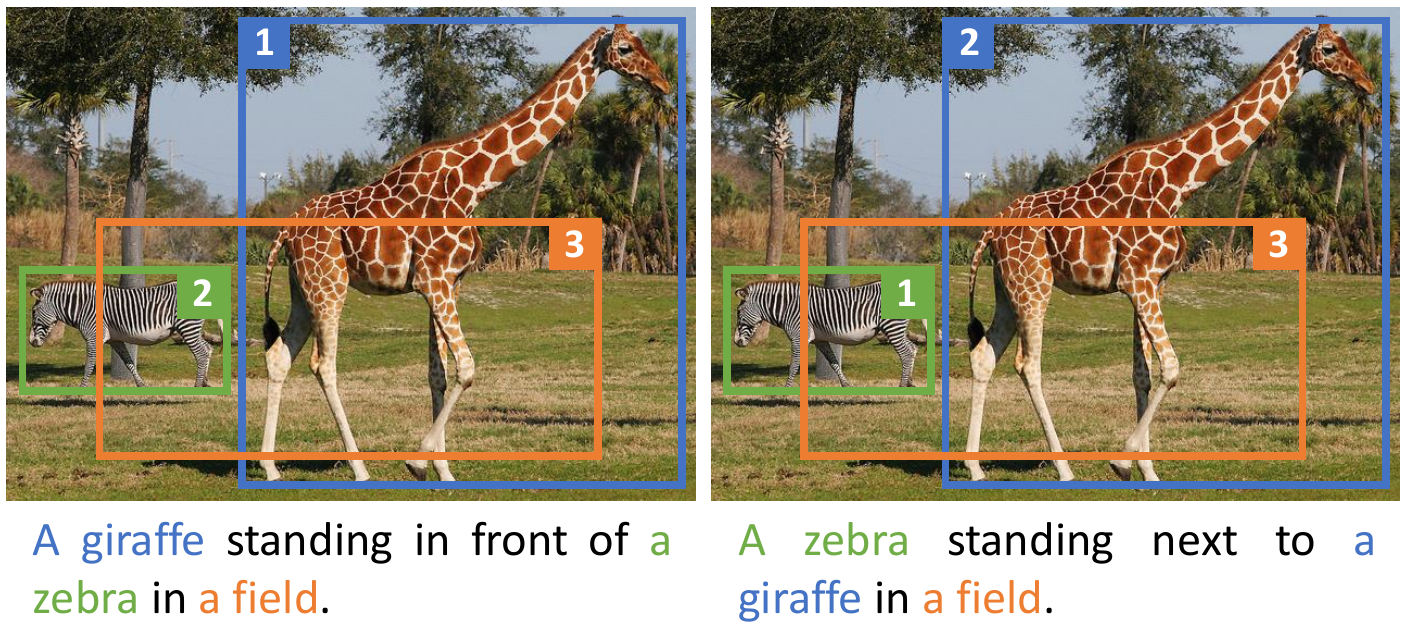} & 
\includegraphics[width=0.33\linewidth,valign=t]{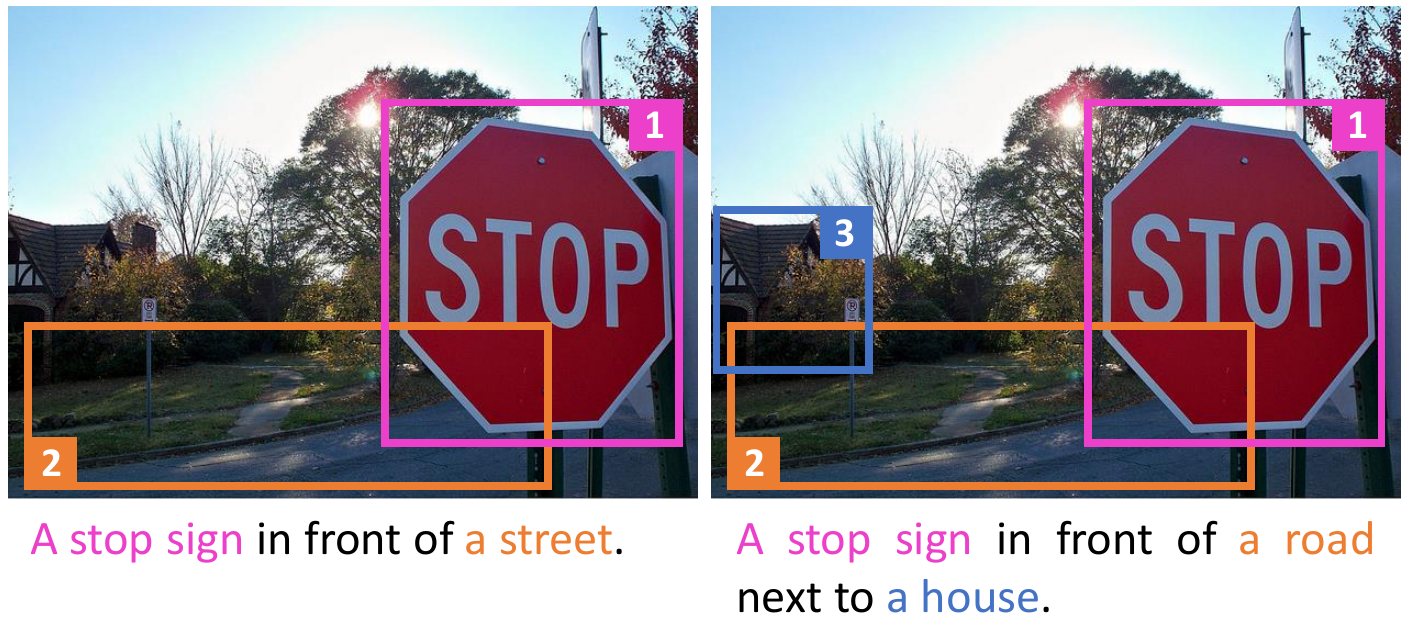} \\
\includegraphics[width=0.33\linewidth,valign=t]{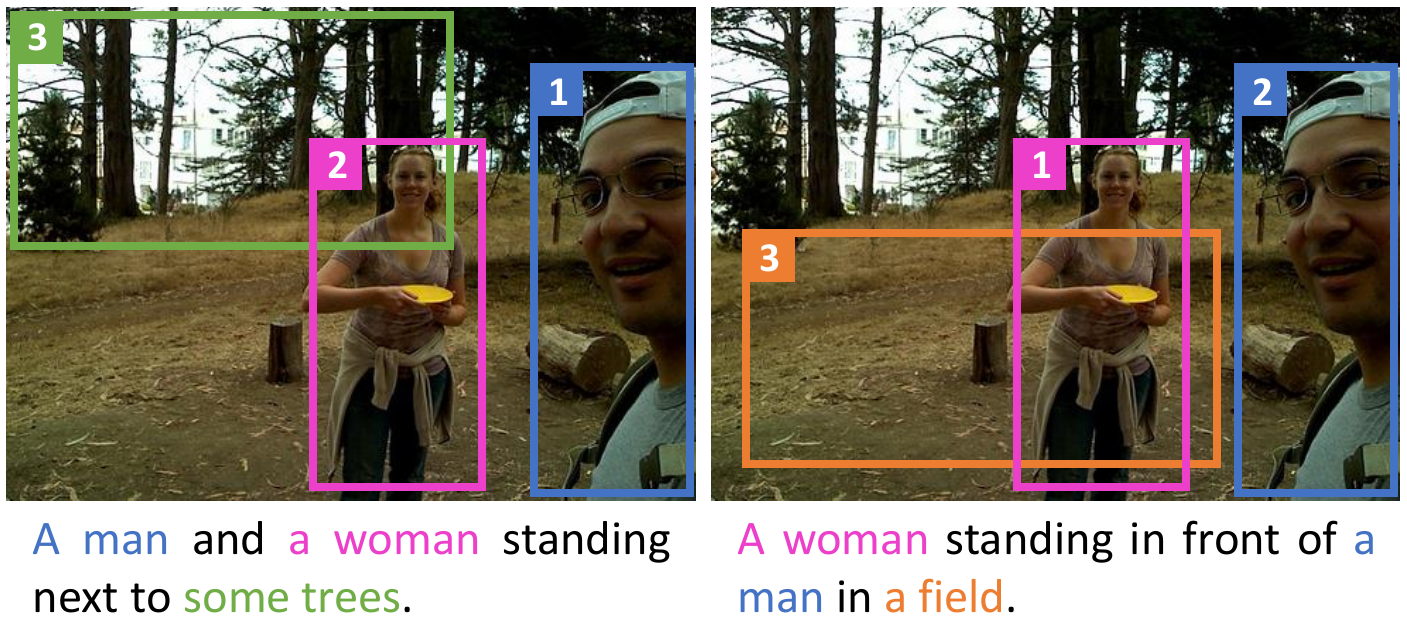} & 
\includegraphics[width=0.33\linewidth,valign=t]{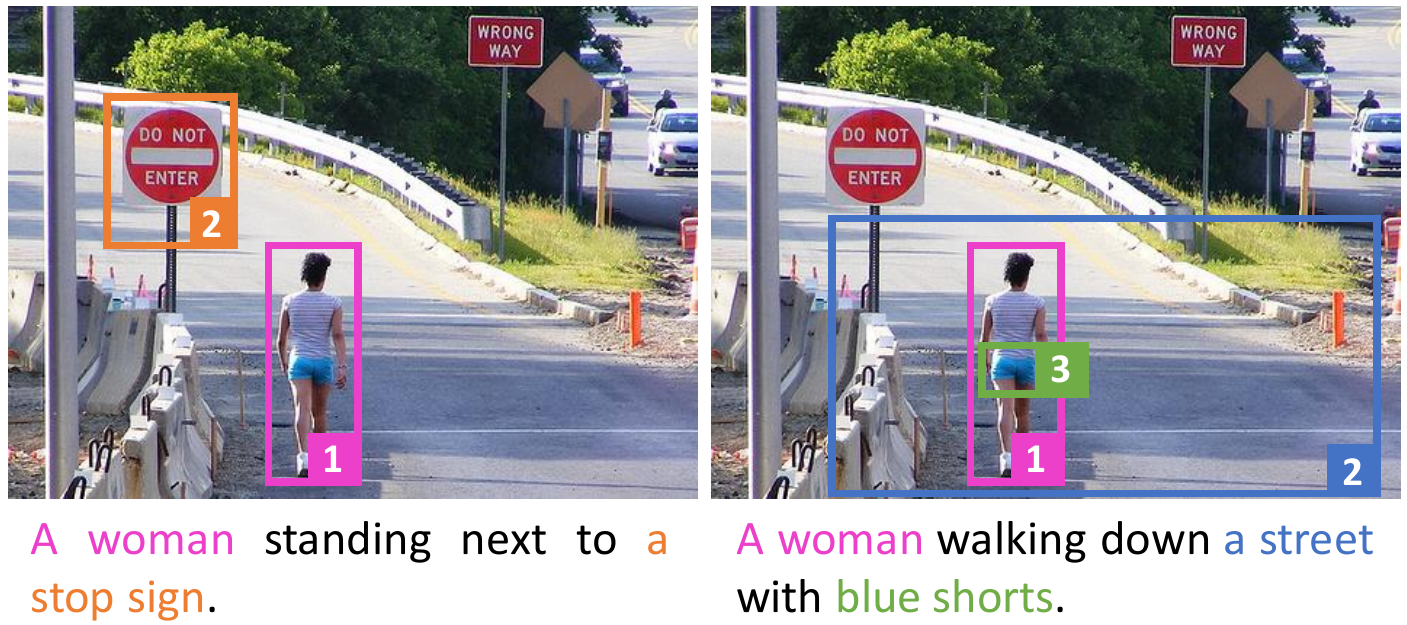} & 
\includegraphics[width=0.33\linewidth,valign=t]{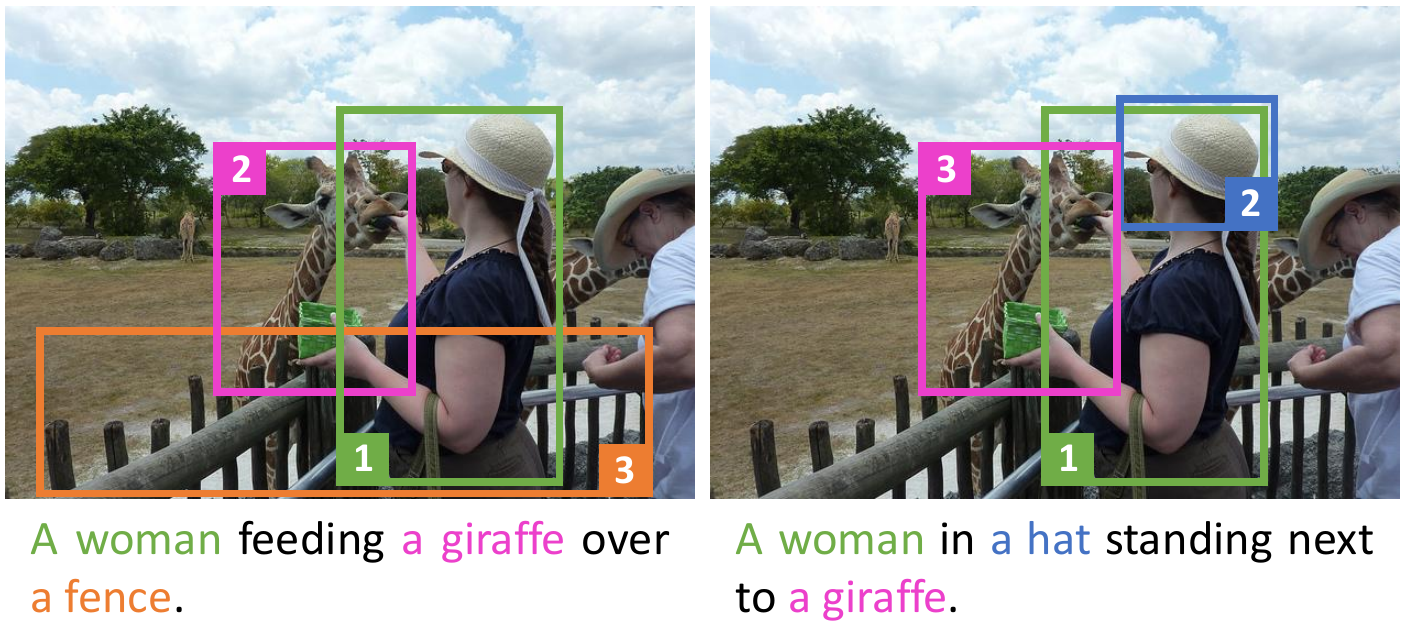} \\
\includegraphics[width=0.33\linewidth,valign=t]{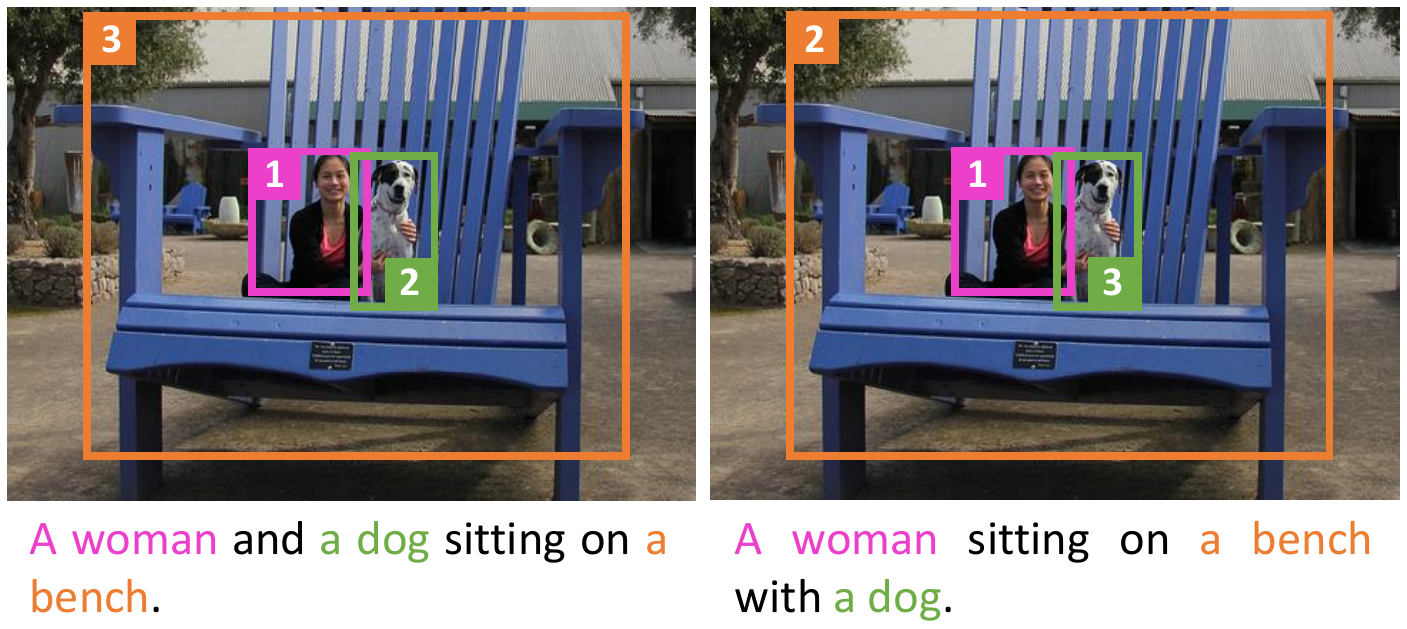} & 
\includegraphics[width=0.33\linewidth,valign=t]{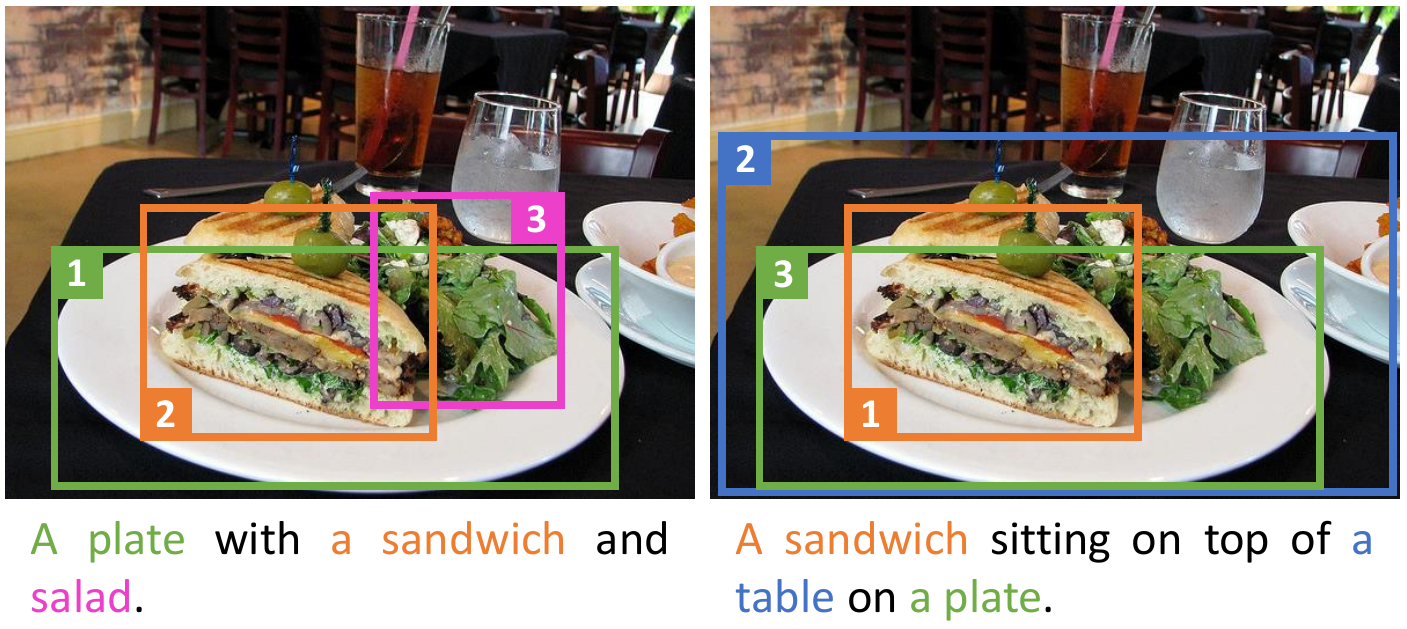} & 
\includegraphics[width=0.33\linewidth,valign=t]{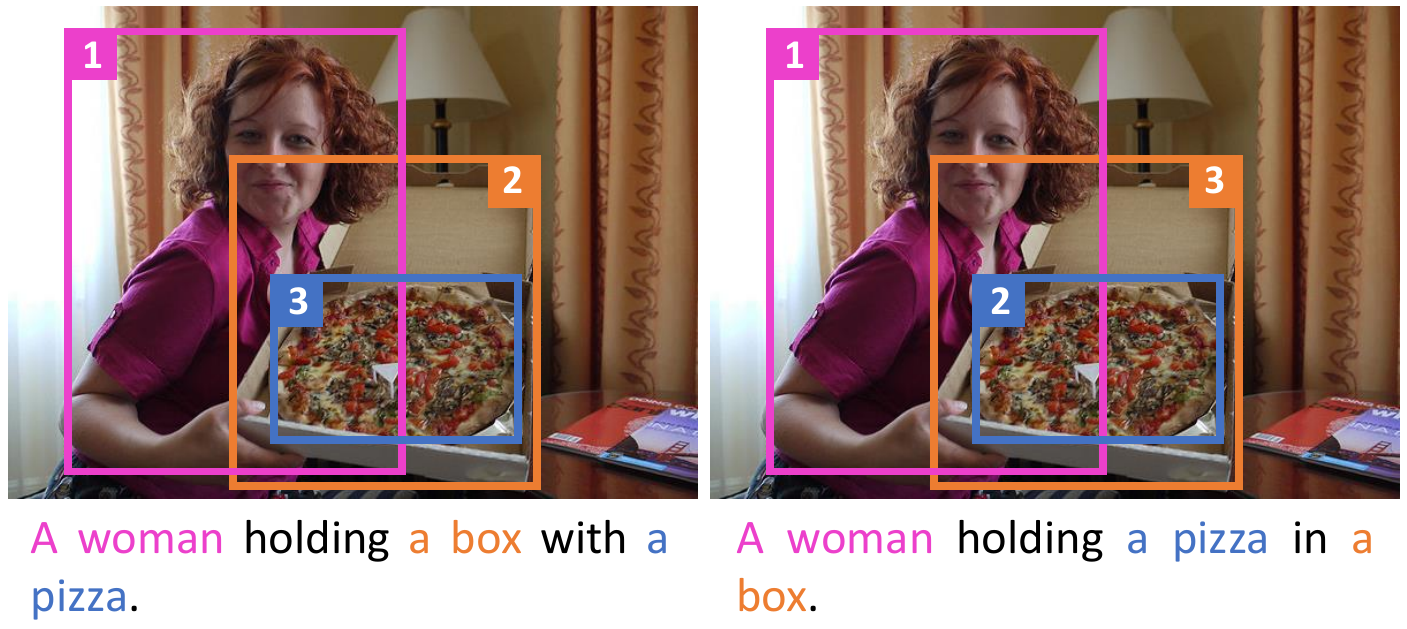} \\
\end{tabular}
\caption{Additional sample results of controllability via a sequence of regions. Different colors and numbers show the control sequence and the associations between chunks and regions.}
\label{fig:supp_sequence_results}
\vspace{2cm}
\end{figure*}

\begin{table*}[t]
\small
\centering
\setlength{\tabcolsep}{.25em}
\resizebox{\linewidth}{!}{
\begin{tabular}{lcccccccccccccccccccc}
\toprule
& \multicolumn{6}{c}{Cross-Entropy Loss} & & \multicolumn{6}{c}{CIDEr Optimization} & & \multicolumn{6}{c}{CIDEr + NW Optimization} \\
\cmidrule{2-7}
\cmidrule{9-14}
\cmidrule{16-21}
Method & B-4 & M & R & C & S & IoU & \:\:& B-4 & M & R & C & S & IoU & \:\:& B-4 & M & R & C & S & IoU \\
\midrule
Controllable LSTM                               & 11.5 & 18.1 & 38.5 & 105.8 & 27.1 & 60.7 & & 12.9 & 18.9 & 40.9 & 122.0 & 28.2 & 62.0 & & 12.9 & 19.3 & 41.3 & 123.4 & 28.7 & 0.642 \\
Controllable Up-Down                            & 17.5 & 23.0 & 46.9 & 160.6 & 38.8 & 69.2 & & 17.7 & 22.9 & 47.3 & 167.6 & 38.7 & 69.4 & & \textbf{18.1} & 23.6 & 48.4 & 170.5 & 40.4 & 71.6 \\
\midrule
\ours \textit{w/} single sentinel               & 16.9 & 22.6 & 46.9 & 159.6 & 40.9 & 70.2 & & 17.9 & 23.7 & 48.7 & 171.1 & 43.5 & 74.4 & & 17.4 & 23.6 & 48.4 & 168.4 & 43.7 & 75.4 \\
\ours \textit{w/o} visual sentinel              & \textbf{17.7} & 23.1 & 47.9 & 166.6 & \textbf{42.1} & 71.3 & & 18.1 & 23.7 & 48.9 & 172.5 & 43.3 & 74.2 & & 17.6 & 23.4 & 48.5 & 168.9 & 43.6 & 75.3 \\
\ours                                           & \textbf{17.7} & \textbf{23.2} & \textbf{48.0} & \textbf{168.3} & \textbf{42.1} & \textbf{71.4} & & \textbf{18.5} & \textbf{23.9} & \textbf{49.0} & \textbf{176.7} & \textbf{43.8} & \textbf{74.5} & & 18.0 & \textbf{23.8} & \textbf{48.9} & \textbf{173.3} & \textbf{44.1} & \textbf{75.5} \\
\bottomrule
\end{tabular}
}
\caption{Controllability via a set of regions, on the test portion of COCO Entities.}
\label{tab:coco_set_results_supp}
\end{table*}

\begin{table*}[t]
\small
\centering
\setlength{\tabcolsep}{.25em}
\resizebox{\linewidth}{!}{
\begin{tabular}{lcccccccccccccccccccc}
\toprule
& \multicolumn{6}{c}{Cross-Entropy Loss} & & \multicolumn{6}{c}{CIDEr Optimization} & & \multicolumn{6}{c}{CIDEr + NW Optimization} \\
\cmidrule{2-7}
\cmidrule{9-14}
\cmidrule{16-21}
Method & B-4 & M & R & C & S & IoU & \:\:& B-4 & M & R & C & S & IoU & \:\:& B-4 & M & R & C & S & IoU \\
\midrule
Controllable LSTM                               & 6.7 & 12.0 & 29.8 & 41.0 & 15.6 & 48.8 & & 6.8 & 12.1 & 30.2 & 45.4 & 15.6 & 49.0 & & 6.4 & 12.5 & 30.2 & 42.9 & 15.6 & 50.8 \\
Controllable Up-Down                            & \textbf{10.1} & \textbf{15.2} & 35.1 & \textbf{68.8} & 21.5 & \textbf{53.6} & & 10.2 & 14.8 & 35.3 & 69.1 & 21.1 & 52.9 & & 10.5 & 15.2 & 35.5 & 69.5 & 21.6 & 54.8 \\
\midrule
\ours \textit{w/} single sentinel               & \textbf{10.1} & \textbf{15.2} & \textbf{35.5} & 67.5 & 21.7 & 52.5 & & 10.1 & 15.3 & 36.1 & 68.9 & 21.7 & 53.5 & & 9.5 & 15.2 & 35.8 & 65.6 & 21.2 & \textbf{55.0} \\
\ours \textit{w/o} visual sentinel              & 9.7 & 14.5 & 34.4 & 63.1 & 21.0 & 52.2 & & 9.9 & 14.7 & 34.8 & 65.5 & 20.8 & 52.9 & & 9.8 & 14.8 & 35.0 & 64.2 & 20.9 & 54.3\\
\ours                                           & 9.9 & 14.9 & 35.3 & 67.3 & \textbf{22.2} & 52.7 & &  \textbf{10.8} &  \textbf{15.7} &  \textbf{36.4} &  \textbf{71.3} &  \textbf{22.0} &  \textbf{53.9} & & \textbf{10.9} & \textbf{15.8} & \textbf{36.2} & \textbf{70.4} & \textbf{21.8} & \textbf{55.0} \\
\bottomrule
\end{tabular}
}
\caption{Controllability via a set of regions, on the test portion of Flickr30K Entities.}
\label{tab:flickr_set_results_supp}
\vspace{-.3cm}
\end{table*}

We observe that the CIDEr+NW fine-tuning approach is effective on all settings, and that our model outperforms by a clear margin the baselines both when controlled via a sequence and when controlled by a scrambled set of regions, regardless of the careful choice of the baselines. The performance of the Controllable LSTM baseline is constantly significantly lower than that of the Controllable Up-Down, thus indicating both the importance of an attention mechanism and that of having a good representation of the control signal. The Controllable Up-Down baseline, however, shows lower performance when compared to our approach, in both sequence- and set-controlled scenarios.

\section{Additional qualitative results}
\begin{figure*}[t]
\centering
\setlength{\tabcolsep}{.08em}
\def\arraystretch{1.3}
\begin{tabular}[t]{ccc}
\includegraphics[width=0.33\linewidth,valign=t]{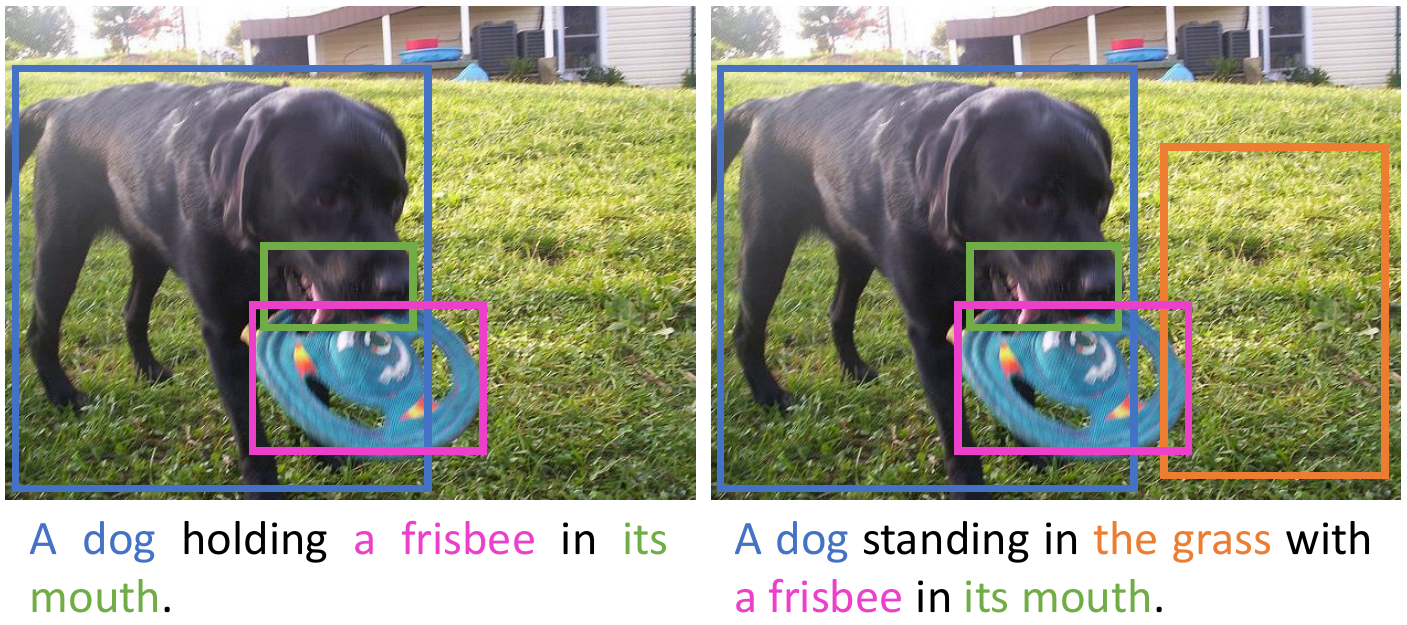} & 
\includegraphics[width=0.33\linewidth,valign=t]{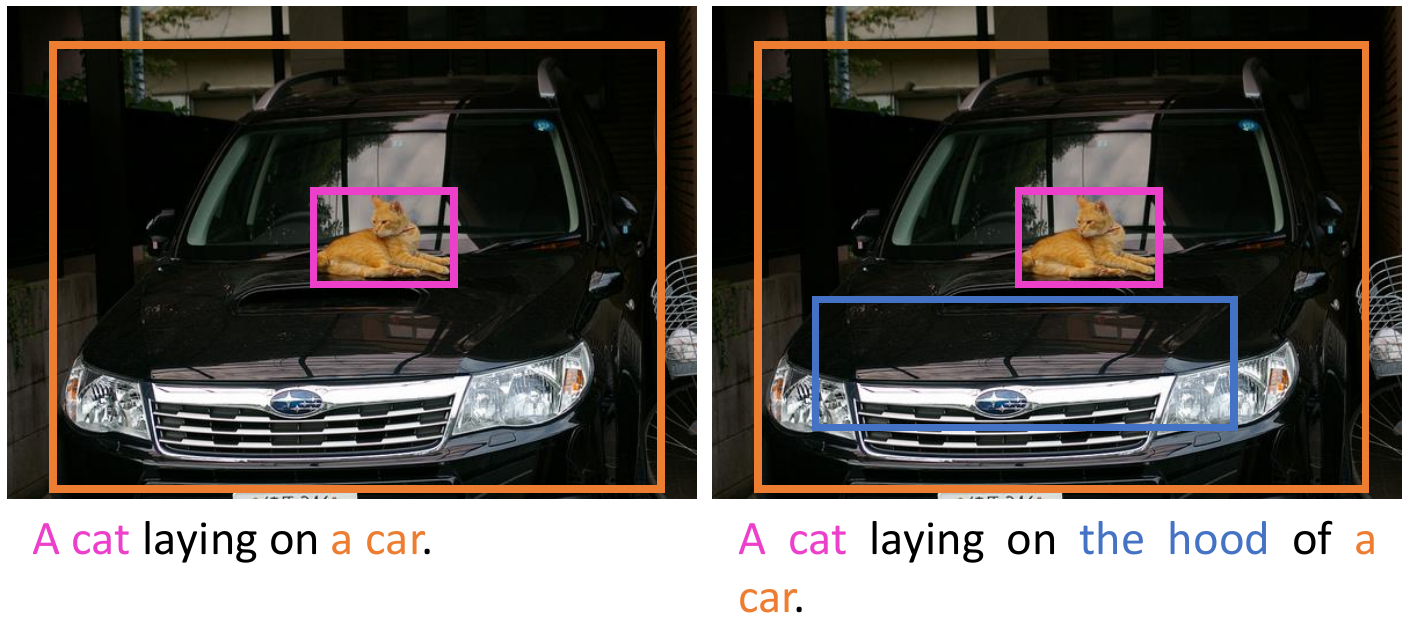} & 
\includegraphics[width=0.33\linewidth,valign=t]{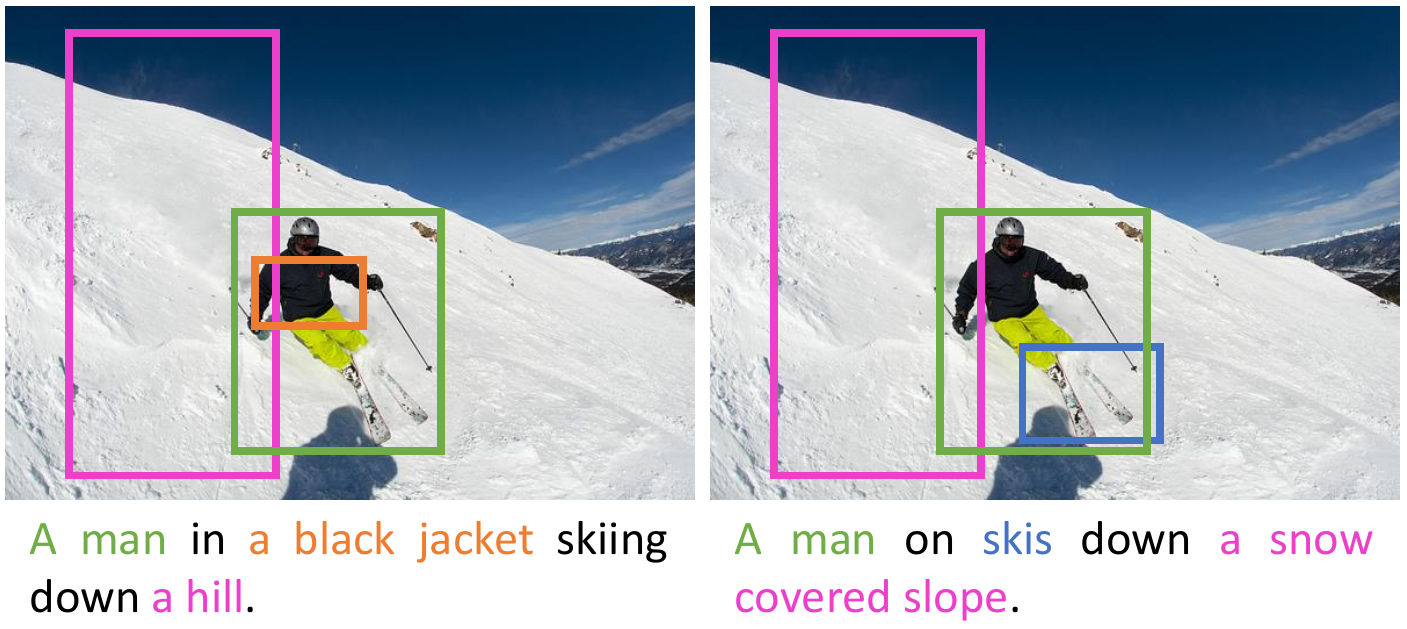} \\
\includegraphics[width=0.33\linewidth,valign=t]{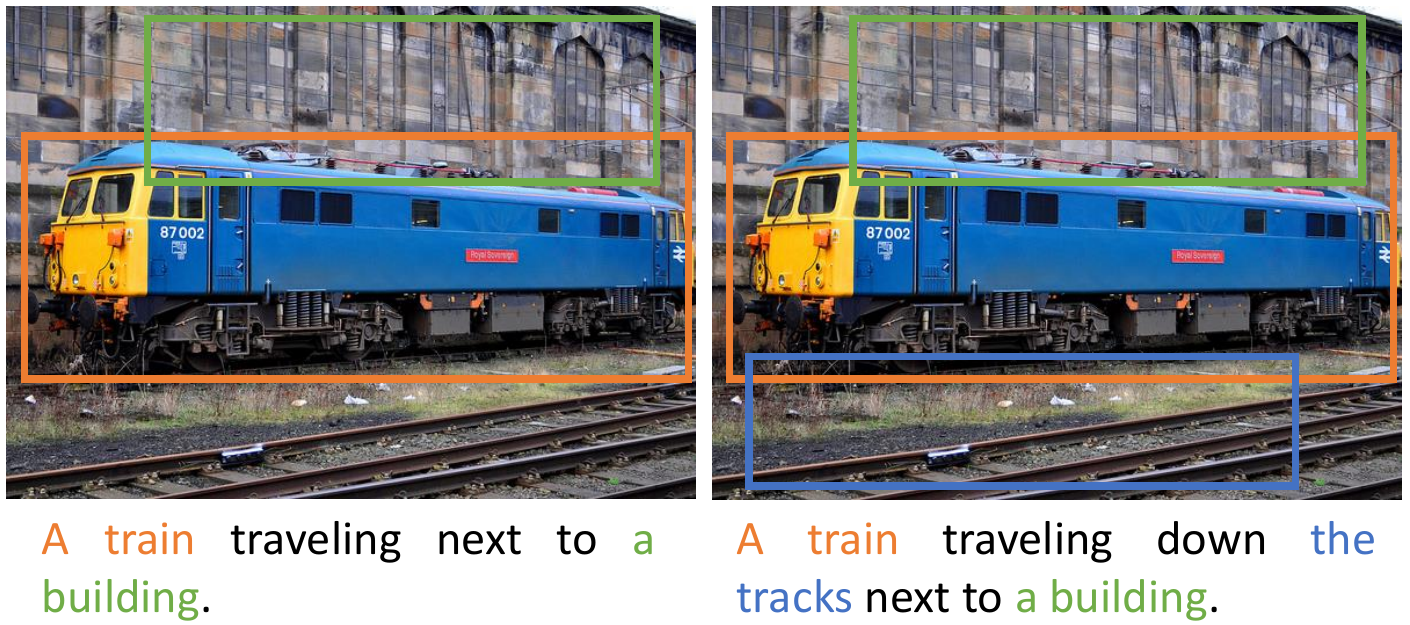} & 
\includegraphics[width=0.33\linewidth,valign=t]{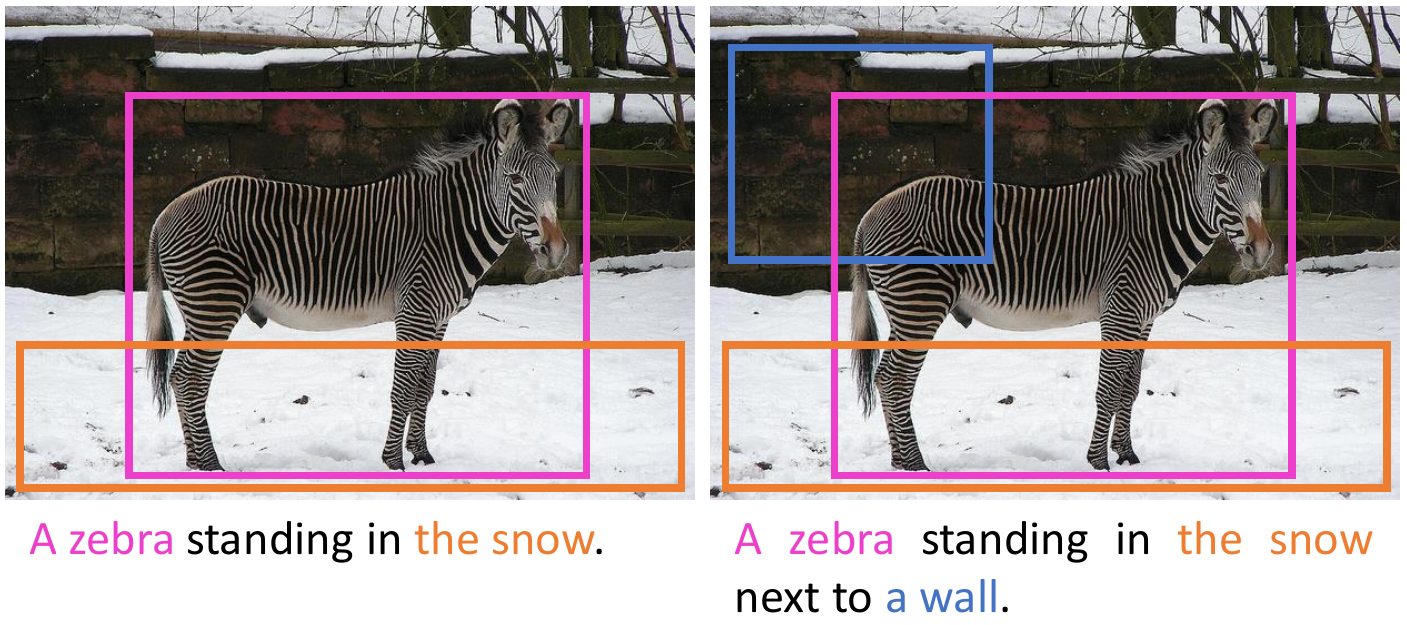} & 
\includegraphics[width=0.33\linewidth,valign=t]{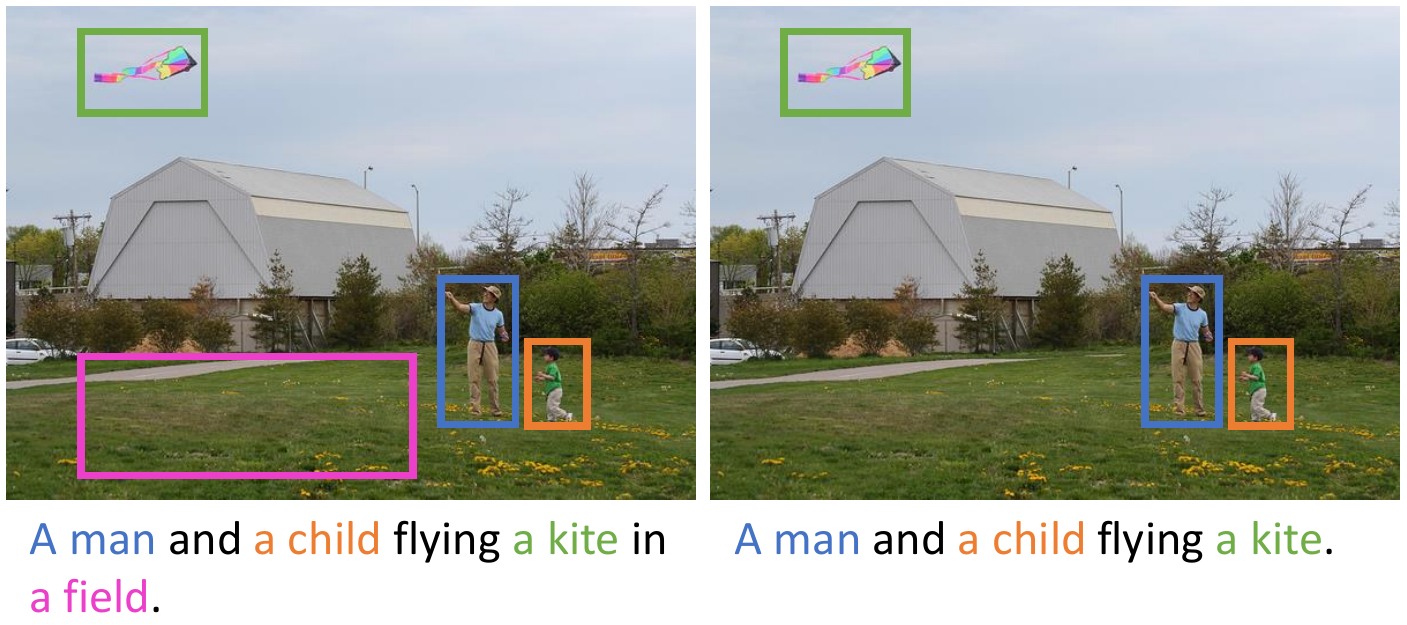} \\
\includegraphics[width=0.33\linewidth,valign=t]{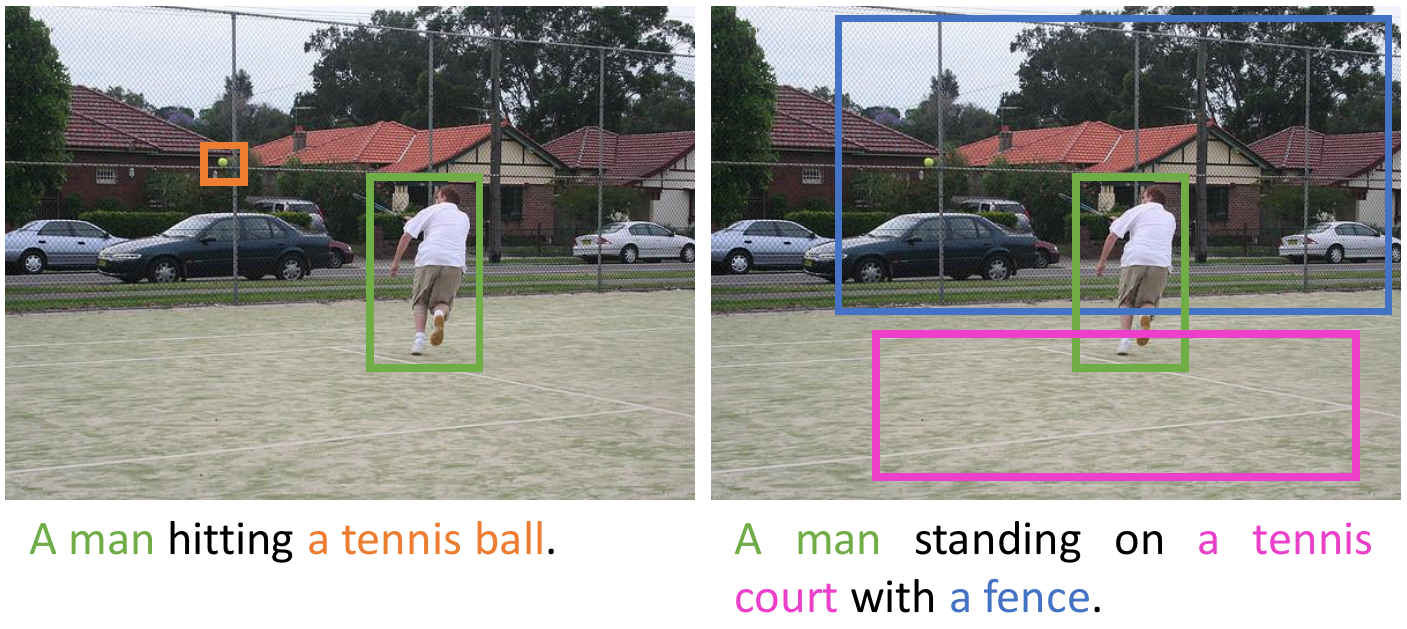} & 
\includegraphics[width=0.33\linewidth,valign=t]{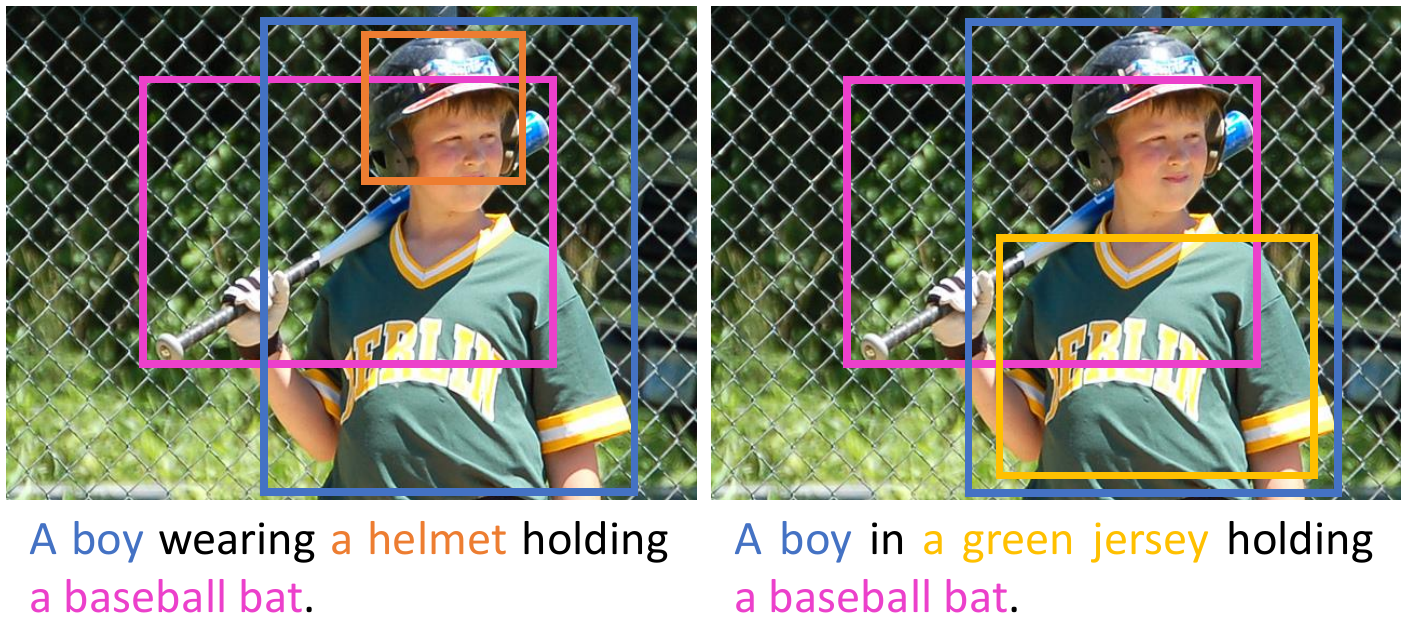} & 
\includegraphics[width=0.33\linewidth,valign=t]{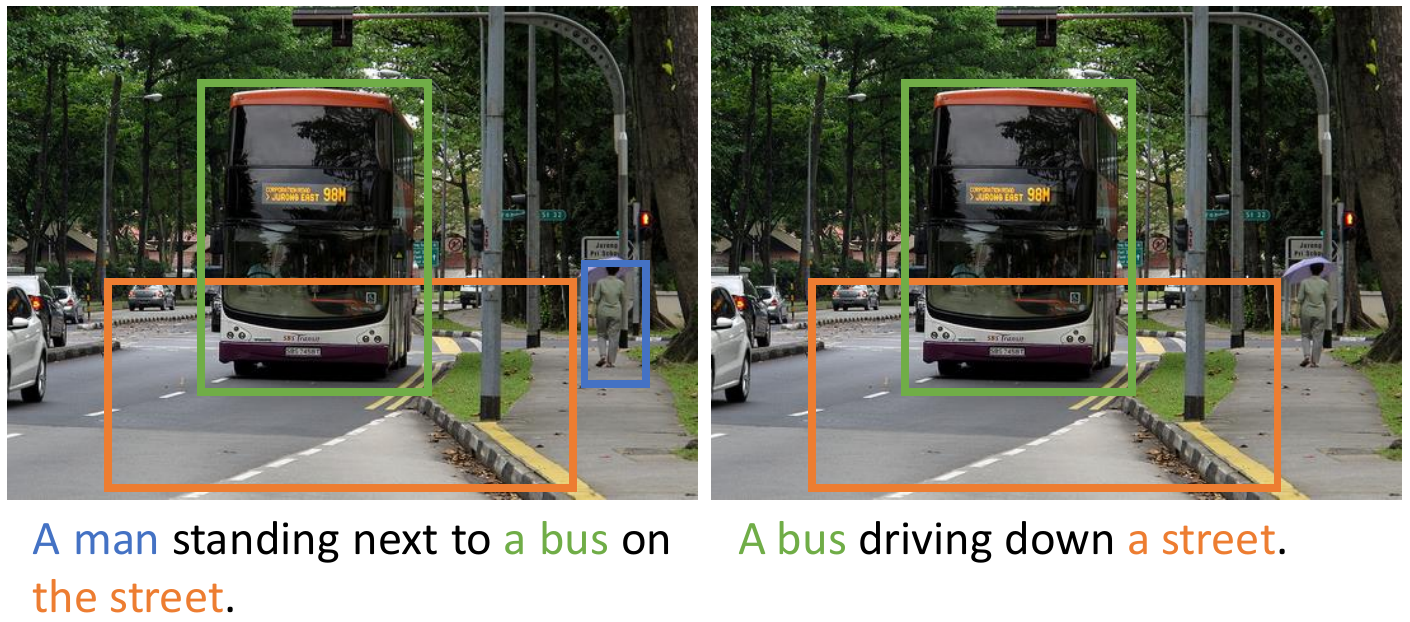} \\
\includegraphics[width=0.33\linewidth,valign=t]{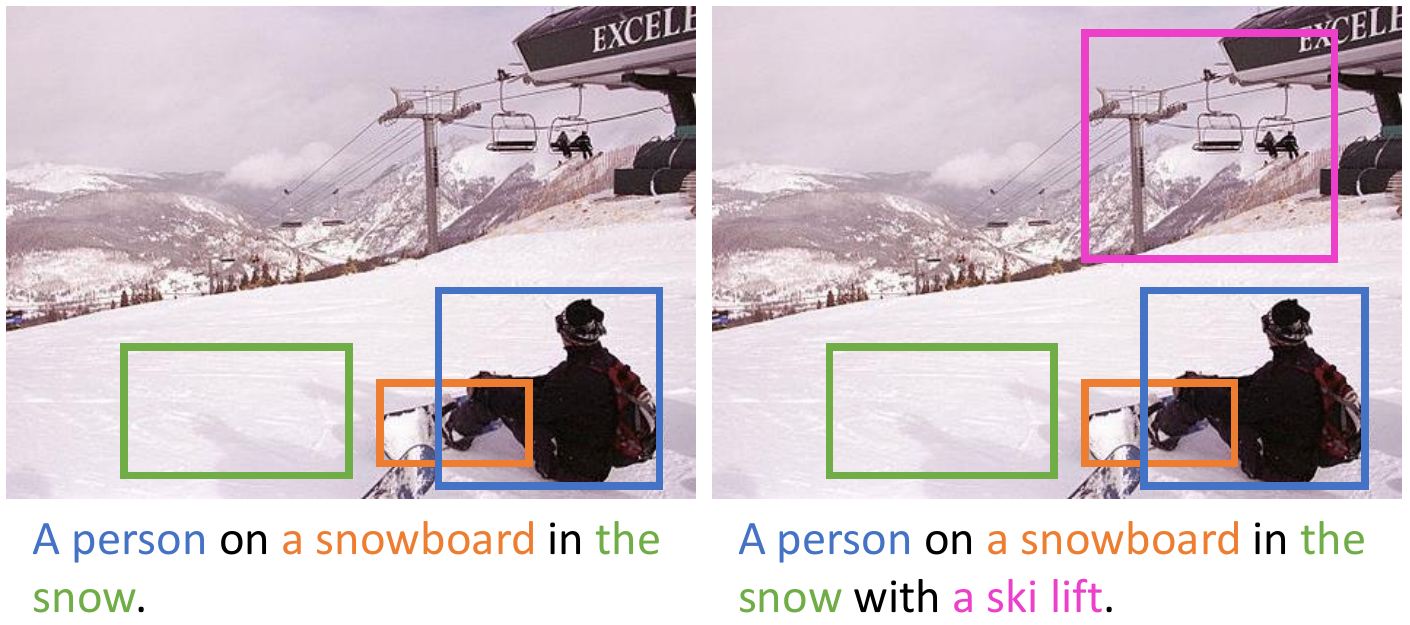} & 
\includegraphics[width=0.33\linewidth,valign=t]{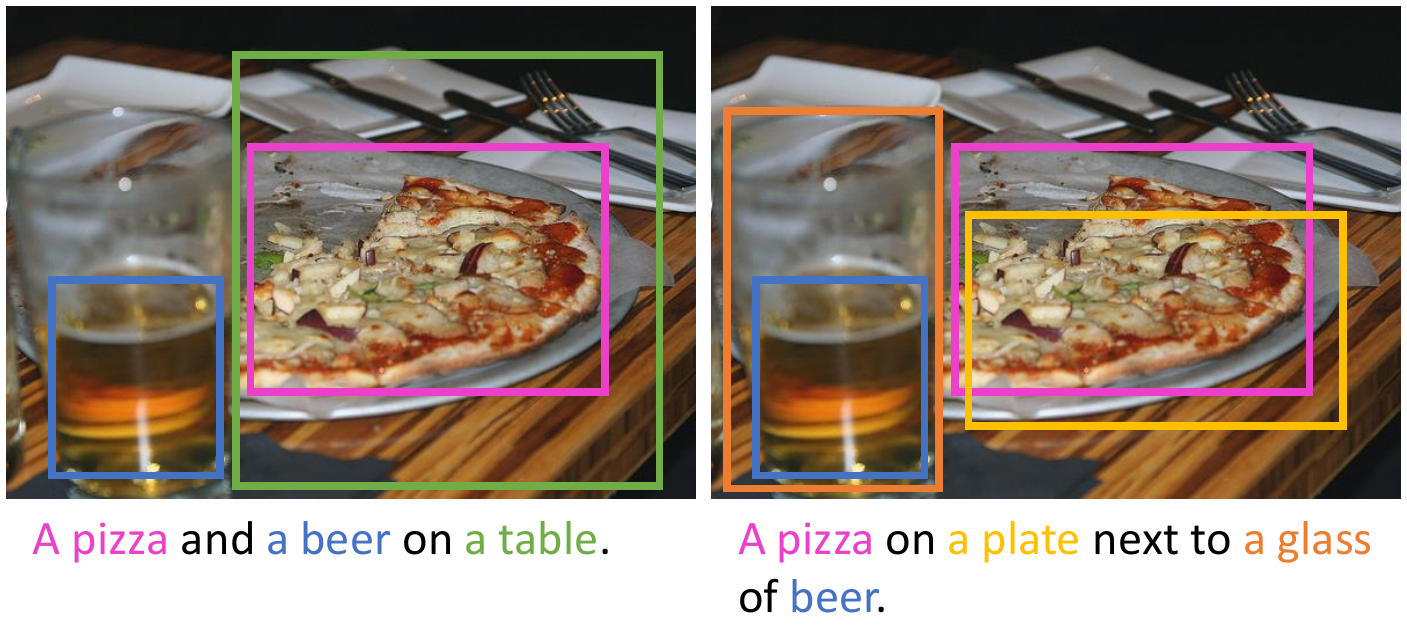} & 
\includegraphics[width=0.33\linewidth,valign=t]{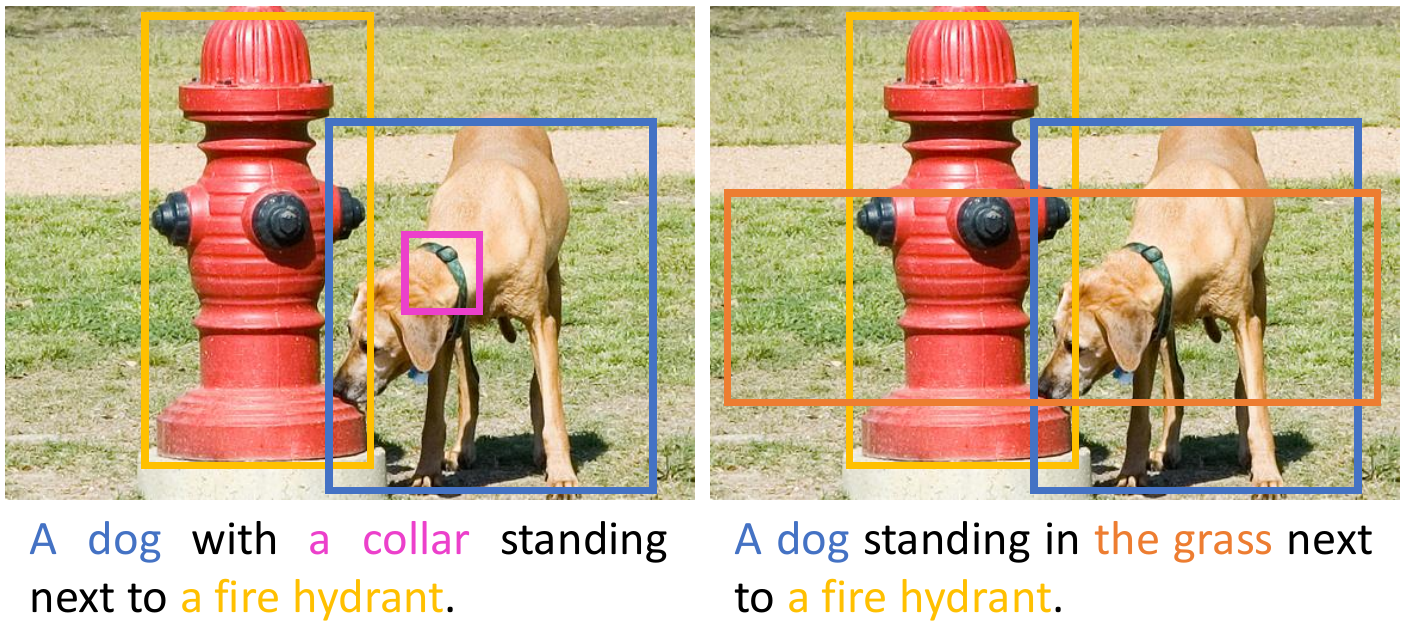} \\
\end{tabular}
\caption{Additional sample results of controllability via a set of regions. Different colors show the control set and the associations between chunks and regions.}
\label{fig:supp_set_results}
\end{figure*}

Finally, Fig.~\ref{fig:supp_sequence_results} and~\ref{fig:supp_set_results} report other qualitative results on COCO Entities. As in the main paper, the same image is reported multiple times with different control inputs: our method generates multiple captions for the same image, and can accurately follow the control input.

\end{document}